\documentclass{article} 
\usepackage{iclr2024_conference,times}

\usepackage{xcolor}
\iclrfinalcopy

\usepackage{natbib}

\usepackage{graphicx}
\usepackage{subcaption}

\usepackage[textsize=scriptsize]{todonotes}
\usepackage{xspace}

\usepackage{algorithm}
\usepackage[noend]{algpseudocode}
\usepackage{booktabs}
\usepackage{bbm}

\usepackage[utf8]{inputenc} 
\usepackage[T1]{fontenc}    
\usepackage{hyperref}       
\usepackage{url}            
\usepackage{booktabs}       
\usepackage{amsfonts}       
\usepackage{nicefrac}       
\usepackage{microtype}      
\usepackage{xcolor}         
\usepackage{xspace}
\usepackage{etoolbox}
\usepackage{enumitem}
\usepackage{amsmath}

\usepackage{amsthm}
\usepackage{tabularx} 

\newcommand{\algname}{\textsc{RAMBO}\xspace}

\newcommand{\denselist}{\itemsep 0pt\topsep-10pt\partopsep-6pt}

\newcommand{\figref}[1]{Fig.~\ref{#1}}

\DeclareMathOperator*{\argmin}{arg\,min}
\DeclareMathOperator*{\argmax}{arg\,max}

\newcommand{\expctover}[2]{\mathbb{E}_{#1}\!\left[#2\right]}

\newcommand{\Unlabeled}{\ensuremath{\mathcal{U}}}

\newcommand{\LabeledSet}{\ensuremath{\mathcal{S}}}

\newcommand{\utilitysample}{\ensuremath{\xi}}

\newtheorem{definition}{Definition}

\newcommand{\unit}{\ensuremath{\mathbbm{1}}}

\newif\iffinal

\iffinal
    
    \newcommand{\zixin}[1]{}
    \newcommand{\ruoxi}[1]{}
    \newcommand{\si}[1]{}
    \newcommand{\siil}[1]{}
    \newcommand{\ruoxiil}[1]{}
    \newcommand{\addpagenumber}{}
\else
    \newcommand{\zixin}[1]  {\todo[fancyline,color=orange]{Zixin: #1}\xspace}
    
    \newcommand{\ruoxi}[1]{\todo[fancyline,color=yellow]{Ruoxi: #1}\xspace}
    \newcommand{\si}[1]{\todo[fancyline,color=cyan]{Si: #1}\xspace}
    \newcommand{\siil}[1]{\textbf{\textcolor{cyan}{[Si: #1]}}}
    \newcommand{\ruoxiil}[1]{\textbf{\textcolor{red}{[Ruoxi: #1]}}}
    \newcommand{\addpagenumber}{\pagenumbering{arabic}}
\fi

\title{Learning to Rank for Active Learning via Multi-Task Bilevel Optimization}
\author{%
        Zixin Ding \\
        The University of Chicago\\
        \And 
        Si Chen \\
        Virginia Tech\\
        \And  
        Ruoxi Jia \\
        Virginia Tech \\
        \And 
        Yuxin Chen\\
        The University of Chicago\\
}

\begin{document}

\maketitle

\begin{abstract}
Active learning is a promising paradigm to reduce the labeling cost by strategically requesting labels to improve model performance. However, existing active learning methods often rely on expensive acquisition function to compute, extensive modeling retraining and multiple rounds of interaction with annotators. To address these limitations, we propose a novel approach for active learning, which aims to select batches of unlabeled instances through a learned surrogate model for data acquisition. A key challenge in this approach is developing an acquisition function that generalizes well, as the history of data, which forms part of the utility function's input, grows over time. Our novel algorithmic contribution is a bilevel multi-task bilevel optimization framework that predicts the relative utility---measured by the validation accuracy---of different training sets, and ensures the learned acquisition function generalizes effectively. For cases where validation accuracy is expensive to evaluate, we introduce efficient interpolation-based surrogate models to estimate the utility function, reducing the evaluation cost. We demonstrate the performance of our approach through extensive experiments on standard active classification benchmarks. By employing our learned utility function, we show significant improvements over traditional techniques, paving the way for more efficient and effective utility maximization in active learning applications. 

\end{abstract}

\addpagenumber


\section{Introduction}


\looseness -1 
Many decision making tasks involve maximization of utility functions \citep{chen2015submodular, jackson2019value}.  
%
Specifically, utility in active learning (AL) can be represented in various forms, such as the expected reduction in error rate 
\citep{mussmann2022active, roy2001toward}, mutual information between the labeled and unlabeled datasets \citep{sourati2016classification, adaimi2019leveraging, lindley1956measure}, or the uncertainty of model predictions \citep{settles2012active, shen2017deep, kossen2022active}. 
However, maximizing utility under budget constraints in active learning is notoriously challenging. It is well-known that determining the optimal set containing maximal information under cardinality constraint is NP-hard \citep{ko1995exact, chen2015sequential}. Moreover, evaluating utility functions can be computationally demanding. 
In active classification, for instance, determining the utility involves retraining the classifier to get validation accuracy. 
The challenge becomes particularly pronounced in deep active learning, where training neural networks can be time consuming, considering the
model cannot be reinitialized from previous rounds of optimization without degrading generalization performance \citep{saran2023streaming}. 
%

Popular approaches for 
active learning often rely on acquisition functions with high \textit{adaptivity} to the environment, in the sense that the selection choices of instances for current round depend on the responses to the labeling requests in all the previous rounds.
This reliance poses major concerns for the deployment of these algorithms to real-world applications, as there could be a substantial delay between requesting labels and receiving feedback. 
For instance, in scientific experiments, feedback from wet-lab or physics experiments can take days or even months to obtain \citep{botu2015adaptive, yang2019machine}. 
This can limit the rounds of interactions with labelers, thus bearing the risk of sampling redundant or less effective training examples within a batch.

Motivated by the above use cases, we study active learning problems with limited adaptivity. The main research question we address is: \textit{How can we develop a robust acquisition criterion for active learning with single round of interaction with annotators within certain budget constraints?}
%
%
Existing active learning approaches often rely on customized utility metrics characterizing the current model's behavior. 
Recent works \citep{ash2019deep, killamsetty2021glister, saran2023streaming, sener2017active} propose to use gradients of the current model based on the \textit{pseudo labels} of the unlabeled data. However, these gradient estimates can be unreliable in a single round active learning setting due to the limited labeled training data. 
The datamodels framework \citep{ilyas2022datamodels} brings to light the linear relationship between training data and model predictions, which seems to be a promising alternative for the design choice of utility model.
It's worth noting that the framework requires \textit{labeled} subsets of training data and studies how the images present in the training set change model predictions. Conversely, we study how to design learning-based acquisition function to map from \textit{unlabeled} instances, or instances without label information to real utility value.

To address these limitations, our approach focuses on enhancing the \textit{robustness} and \textit{generalizability} of active learning, especially when working with deep neural networks. Given the variability in deep learning models due to different initializations, hyperparameters, network architectures and training procedures \citep{jiang2021assessing, d2022underspecification,zhong2021larger}, the one-shot estimate of validation accuracy can be highly stochastic, and thus we turn to the idea of \textit{ranking} as a strategy to mitigate the inherent uncertainty. In a nutshell, instead of learning a predictor for validation accuracy, we shift the perspective towards comparing which subset of training data would provide more useful information for generalization. Concretely, we aim to predict the relative utility value of equal size subset of training data via a novel variant of RankNet \citep{burges2005learning} by inserting set-based neural network architecture to extend comparisons between pairs of examples to pairs of sets. 
By incrementally collecting subsets of labeled pool as the progression of the active learner, we learn a 
batch acquisition function, referred as \textit{utility function}, mapping subsets to utility value. 

\looseness -1 To make our approach generalizable to growing size of collected subsets of labeled data, 
we categorize samples based on the size of inputs and employ bilevel training to account for the growing training history. Additionally, we introduce a multi-task learning framework that uses the optimal transport distance \citep{alvarez2020geometric} between the current labeled data and validation set as a supplementary loss, 
regularizing the utility model to be more closely aligned with the validation distribution, while being oblivious to training dynamics of the underlying classification model.  
In summary, 

\begin{figure*}[t]%
\centering
  \includegraphics[width=\textwidth]{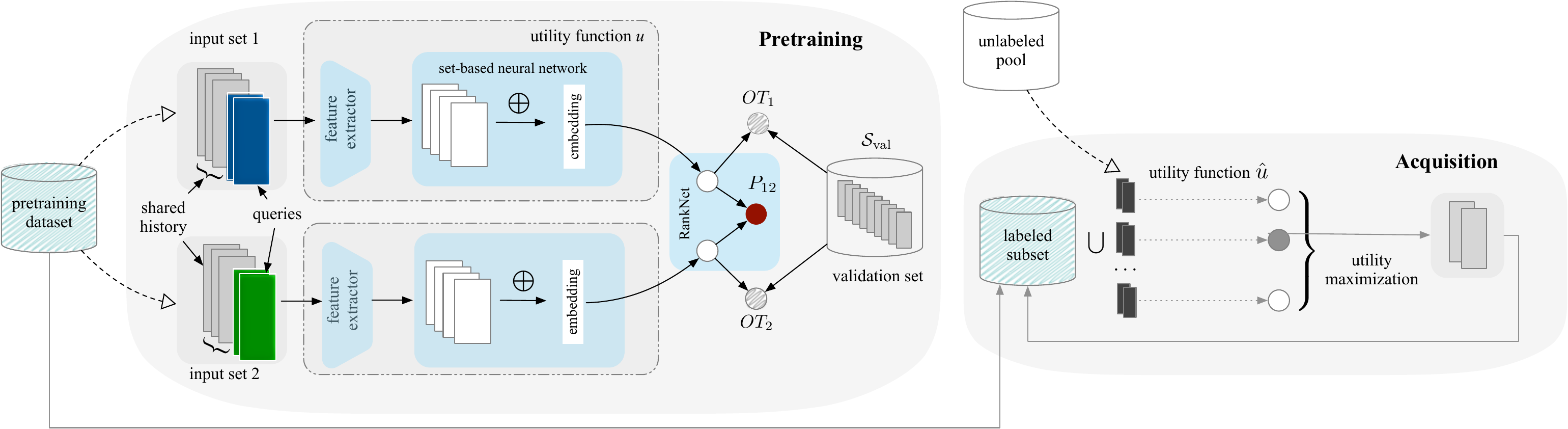}
  \vspace{-2mm}
  \caption{Overview of the \algname algorithm. In the pretraining stage, we learn a RankNet over pairs of utility samples via multi-task bilevel optimization; in the acquisition stage, we follow the learned utility function to iteratively query data points in minibatches. Details of the algorithm are provided in Section~\ref{DUAL_MAX}.}\label{fig:overview}
  \vspace{-2mm}
\end{figure*}

\vspace{-2mm}
\begin{itemize}[leftmargin=*] \denselist
\item We propose a novel single-round active learning approach called \algname (\underline{R}anking-based \underline{A}ctive learning via \underline{M}ultitask \underline{B}ilevel \underline{O}ptimization) that addresses the limitations of existing methods reliant on expensive acquisition functions or overly generic heuristics. Our algorithm is summarized in \figref{fig:overview}.
\item We introduce a bi-level learning algorithm to enhance validation performance, enabling the learning of generalizable utility functions as the history of data grows over time.
\item We employ interpolation-based techniques to augment utility samples (defined in Section~\ref{sec:framework}), 
refining utility model estimation and reducing the requests for groundtruth utility samples during the pretraining stage.
\item 
We incorporate a multi-task learning approach , leveraging the optimal transport distance between the labeled dataset and validation set as a regulatory loss, guiding the behavior of our RankNet.
\item We conduct extensive experiments on various image classification tasks, demonstrating the effectiveness of our proposed approach. Our method also offers a promising alternative for maximizing data utility under budget constraints, with potential applications in a wide range of machine learning tasks.
\end{itemize}
\vspace{-1mm}

\section{Related Work} 

%
\paragraph{Utility model learning}
\citet{konyushkova2017learning} present one can train a regressor that predicts the expected error reduction for a candidate sample during data acquisition stage in active learning. \citet{coleman2019selection} demonstrate that even simple models, when retrained, can serve as effective proxies for acquisition functions. Surrogate models have been leveraged by \citet{li2021active} to approximate the distribution of labels and unobserved features, while \citet{kossen2022active} utilize them to assist data acquisition, albeit with predefined acquisition criteria.
Recent work \citep{ilyas2022datamodels} assumes that model performance is linearly \textit{predictable} via extensive training of utility models.
In contrast, our method use ranking-based neural networks to directly \textit{learn} a data acquisition function predicting which subset would yield higher utility value given a pair of equal size of subset training data. Contrary to \citet{ilyas2022datamodels}, we learn utility function by sampling from various sizes of subsets rather than fixing subset sizes. 
\paragraph{Planning-based vs learning-based AL strategies}
Classical AL has explicitly defined 
query strategy including
uncertainty sampling \citep{settles2012active, shen2017deep, gal2017deep} and diversity sampling \citep{sener2017active, yehuda2022active} or their combined approaches \citep{xie2022towards, citovsky2021batch}. One popular acquisition function with deep neural networks is \citet{ash2019deep} which selects a subset based on diverse gradient embedding obtained by hypothesized samples, trading off between sample diversity and model uncertainty. Similarly, \citet{sener2017active} consider diversity sampling on the penultimate layer representation induced by the current state of the classifier. 
Meanwhile, there is a long line of work on \textit{learning-based} acquisition function 
\citep{learntoactivelearning, learnalgoforactivelearning, oneroundactivelearn, sinha2019variational, yan2022budget, yoo2019learning, li2021active, killamsetty2021glister}. For instance, \citet{killamsetty2021glister} cast designing acquisition function into bi-level optimization framework for jointly optimizing model parameters for both training and validation loss. \citet{borsos2021semi} leverage bi-level optimization to design learning-based batch acquisition function in the fashion of semi-supervised learning. \citet{yoo2019learning} adopt the idea of ranking the predicted classifier loss in comparing two instances as ``loss prediction module'', querying instances that the classifier is likely to predict wrong, and learn it to predict target losses of unlabeled inputs. 
We draw inspirations from \citet{killamsetty2021glister, yoo2019learning} by leveraging bi-level training as a subroutine for learning generalizable utility model incorporating growing history of labeled pool and reducing utility maximization problem by selecting highest ranked batch of unlabeled instances.
\vspace{-1mm}
\paragraph{Learning to rank}
Ranking techniques have been foundational in fields such as 
information retrieval \citep{liu2009learning} and 
recommendation systems \citep{karatzoglou2013learning,li2022learning}. Motivated by \citet{yoo2019learning, li2021learning}, we shift from the traditional approach of directly learning cross-entropy loss on unlabeled instances 
to ranking the utility for paired subsets of data.
While both works \citep{yoo2019learning, li2021learning} view ranking predicted losses as an uncertainty measure, our methodology centers on gauging the utility of labeled data subsets, with the utility being the validation accuracy post-training.
To the best of our knowledge, our method is the first to incorporate the idea of ranking between pairs of subsets and link it directly to performance of the learning algorithm on the validation set. Our unique contribution lies in introducing such a ranking mechanism and integrating it under the RankNet \citep{burges2005learning} framework, in tandem with plugging in optimal transport distance under a multitask learning scenario. 

\section{Problem Statement}\label{ProblemStatement}

Consider a ground set of data points \(\mathcal{X}\) with a ground truth labeling function $f^*: \mathcal{X} \rightarrow \mathcal{Y}$. The active learning problem in our study unfolds in a two-stage protocol: a \textit{pretraining stage} and an \textit{acquisition stage}. In the pretraining stage, we are given an initial pool of data points; during the acquisition stage, we proceed to actively select a set of new examples to label all at once. We denote the initial pretraining (labeled) set by \(\LabeledSet_0\) 
with $\LabeledSet_0 \subseteq \mathcal{X}$ and $|\LabeledSet_0|=k$, and denote the labeled set after the acquisition stage by \(\LabeledSet_1\) with \(|\LabeledSet_1|=k+B\) where \(B\) represents the labeling budget. The unlabeled set at initiation and after acquisition is represented as \(\Unlabeled_0\) and \(\Unlabeled_1\) respectively.

The groundtruth utility function is defined as \(u: 2^{\mathcal{X}} \rightarrow \mathbb{R}\), where $u(\mathcal{\utilitysample})$ quantifies the utility of a subset \(\mathcal{\utilitysample} \subseteq \mathcal{X}\) by evaluating the validation accuracy of the classifier \(f\) induced by the (labeled) data in \(\mathcal{\utilitysample}\). Our goal is to find the optimal subset \(\LabeledSet_1^*\) such that \(f\), trained on it achieves maximal validation accuracy, i.e., optimizes the utility function \(u\):
\begin{align}
    \LabeledSet_1^* \in \argmax_{\LabeledSet_0 \subseteq \LabeledSet_1 \subseteq \mathcal{X}, |\LabeledSet_1 \setminus \LabeledSet_0| = B} u(\LabeledSet_1)
\end{align}
Here, $u(\LabeledSet_1) = \expctover{x}{\unit(f(x) \neq f^*(x))}$ for classification tasks, and can be estimated by the error rate of the resulting $f$ on a validation set $\mathcal{S}_{\text{val}} \subseteq \mathcal{X}$. 

The prime challenge lies in learning the utility function \(u\) that usually requires labeled data to be computed accurately, under the practical constraints of limited labeling budget. We emphasize that the instances are selected \textit{non-adaptively} in the acquisition stage, i.e., our selection of instances do not depend on the label of previous selected instances in the acquisition process. Our goal is to 
devise an acquisition strategy that proficiently selects a data subset for labeling which maximally improves the model's predictive performance.

\section{Methodology} \label{DUAL_MAX}
We introduce our algorithm, \algname, following the two-stage learning protocol described previously. In a nutshell, \algname (1) collects training samples for utility model pretraining, and (2) greedily selects the batch with the maximal predicted utility value from one to total batches $t$ in the acquisition stage. We divide the pretraining stage into $\tau_{1}$ iterations and acquisition stage into $\tau_{2}$ iterations with mini-batch size $b$ for each iteration. More concretely, 
we instantiate 
\algname into following building blocks: a) Develop a set-based multitask neural network model $\hat{u}$ as surrogate model for pretraining; b) define the loss function for the utility model $\hat{u}$; c) sample a collection of subsets $\{(\utilitysample, u(\utilitysample))\}_{i} \subseteq \LabeledSet_{0}$ where $i \in [1, \tau_{1}]$ as a growing labeled set up to $\LabeledSet_{0}$ for training $\hat{u}$; d) update the set based model $\hat{u}$ per iteration of the pretraining stage; e) greedily follow the learned utility model $\hat{u}$ in the acquisition stage.

\subsection{A Two-Stage Active Learning Framework}\label{sec:framework}
We now explicitly introduce this framework and apply it to classification tasks. We will unravel a)-d) above and discuss each relevant aspect respectively:

\textbf{a) What surrogate models $\hat{u}$ should we use?} \label{surrogatemodel}
Similar to \cite{ilyas2022datamodels}, by parametrizing a surrogate model with training samples, we transform the surrogate model construction into supervised learning task (See Definition \ref{UtilityModel}). 
In our context, the training samples $\utilitysample$ are subsets of pretraining set $\LabeledSet_{0}$ and the utility value is $u(\utilitysample)$. Throughout this work, we refer the pairs $(\utilitysample, u(\utilitysample))$ as \textit{utility samples}. It is appealing to adopt their linearity assumption into AL setting due to strong theoretical footing \citep{saunshi2022understanding} and simplicity in model architectures. Nevertheless, to improve efficacy of model prediction in acquisition stage and avoid extensive model retraining as \citet{ilyas2022datamodels}, we hence prefer more complex architectures for modeling interaction between elements within each utility sample. One natural candidate for $\hat{u}$ is set-based neural networks due to their strong expressive power (i.e., Set Transformer \citep{lee2019set} or Deep Sets \citep{zaheer2017deep}). Denote the general set-based neural network(NN) as 
\begin{align}
\label{DeepSets}
    \text{net}(\utilitysample) &=\text{net}(x_{1}, ..., x_{a}) 
    = \rho(\text{pool}(\{\phi(x_{1}), ... \phi(x_{a})\}) \nonumber
\end{align}
where $\{ x_{i} \}_{i = 1}^{a}$ represents a single utility sample $\utilitysample$ with size $a$ and $\phi, \rho$ is the feature extractor and regressor for the set-based NN itself.

In experiments, we find solely inputting data contents to set-based NN is not robust enough for model prediction. \citet{alvarez2020geometric} introduce the notion of geometric distance via optimal transport (OT) between two datasets and \citet{just2023lava} extend it as a learning-agnostic proxy for measuring model performance on $\LabeledSet_{val}$. The celebrated success of OT distance in predicting validation set accuracy \citep{just2023lava} enables us to cast the groundtruth OT distance between utility samples and validation set \citep{alvarez2020geometric} as a supervision signal for extending our model to multitask set-based NN.


\begin{definition}[Surrogate Utility Model]\label{UtilityModel}
    Let $\mathcal{X}$ be the instance domain, and $\utilitysample$ be any sampled subset drawn from distribution $\mathcal{D}$ over $\mathcal{X}$. A \textit{surrogate utility model} 
    $\hat{u} (\utilitysample)$ 
    is a set function mapping from $2^\mathcal{X} \rightarrow \mathbb{R}$, 
    optimized to predict the true utility $u(\utilitysample)$ on a training set $\utilitysample \sim \mathcal{D}$:
    \begin{align}
        \hat{u}=\argmin_{\tilde{u}_w}\mathbb{\hat{E}}_{\utilitysample \sim \mathcal{D}}[\mathcal{L}(\tilde{u}_w(\utilitysample), u(\utilitysample)]
    \end{align}
where $\mathcal{L}(\cdot, \cdot)$ denotes the loss function, and $\tilde{u}_w$ denotes a parametric set function defined to approximate $u$.
\end{definition}

    

\textbf{b) What loss function should we minimize?} 
One natural choice is to directly minimize the MSE(mean square error) of predicted and true utility value as $\mathcal{L} = (\hat{u} - u)^{2}$. However, the evaluation of validation accuracy is non-deterministic (thus stochastic) due to the aleatoric uncertainty of the classifier itself.
While the simplistic way is to train a deep neural network to approximate the utility value in regression fashion and minimize the MSE, we fail to learn a good utility model by regressing validation accuracy on set of utility samples (See Section~\ref{Three Design Choices} for ablation study on casting utility model as regression network). A natural way to design our loss function lies in the idea of pairwise ranking. \cite{yoo2019learning} introduce a loss prediction module to predict the classifier loss on single data point and handicraft the loss function for predicting the classifier loss in pairwise ranking fashion. For a minibatch samples with size $d$, \citet{yoo2019learning} divide it into $d/2$ pairs and rank the differences between each pair of predicted and groundtruth losses to discard the overall loss scale. Extending the idea of ranking classification loss between pairs of data point to rank the utility value, we incorporate the classical RankNet \citep{burges2005learning} structure to rank between pairs of equal size utility samples with OT distance as a regularizer in the final loss.

\begin{definition}[Ranking Loss]
    Given $\mathcal{X}$, and let $\utilitysample_{1}, \utilitysample_{2}$ be two sampled subset drawn from distribution $\mathcal{D}$ over $\mathcal{X}$ with equal size $d$. Denote the utility value (validation accuracy) of $\utilitysample_{1}$ as $u_{1}$ and the utility value of $\utilitysample_{2}$ as $u_{2}$. W.l.o.g. suppose $u_{1} > u_{2}$, $u_{12} = u_{1} - u_{2}$. Specifically, $u_{1} > u_{2}$ is taken to mean that the surrogate utility model $\hat{u}$ asserts that $\utilitysample_{1} \rhd \utilitysample_{2}$. Denote the modeled posterior $P(u_{1} \rhd u_{2})$ by $P_{12}$, and let $\bar{P}_{12}$ be the desired target values for those posteriors. The Binary Cross Entropy(BCE) loss for pair $(\utilitysample_{1}, \utilitysample_{2})$ is written as 
    \begin{align}
\mathcal{L}_{\text{Rank}_{12}} 
        = - \bar{P}_{12}\log P_{12} - (1 - \bar{P}_{12}) \log(1 - P_{12}).\nonumber
    \end{align}
\end{definition}
With this metric in hand, we shall guide $\hat{u}$ to learn the principled signal ties to validation set accuracy and ignore the shifting distribution between labeled data and $\LabeledSet_{val}$ in the acquisition stage (See Definition~\ref{otloss}). Even though OT distance can be approximated in near-linear time complexity \citep{altschuler2017near}, our goal is to predict which subsamples of training data would yield the highest validation set accuracy rather than approximating OT distance itself. To bypass the computational infeasibility within the limited timeframe, we use the OT distance as a groundtruth supervision signal to regularize the $\hat{u}$ in pretraining rather than serving as an input to $\hat{u}$. We show the efficacy of incorporating OT Distance Loss in Section~\ref{Three Design Choices}.
\begin{definition}[OT Distance Loss]
\label{otloss}
    Given two utility samples $\utilitysample_{1}$, $\utilitysample_{2}$ and its corresponding ground truth OT distance value as $OT_{1}$, $OT_{2}$ and the predicted values as $\hat{OT}_{1}$ and $\hat{OT}_{2}$. The loss is defined as 
    \begin{align}
\mathcal{L}_{\text{OT}} = & \lambda_{1} (\hat{OT}_{1} - OT_{1})^{2} + \lambda_{2} (\hat{OT}_{2} - OT_{2})^{2} \nonumber \\
    & - \lambda_{3} (\min(\hat{OT}_{1}, 0) + \min(\hat{OT}_{2}, 0)) \nonumber
\end{align}
where $\lambda_{1}, \lambda_{2}, \lambda_{3}$ are hyperparameters. The first two terms are mean squared error for OT distances and the third terms are positive constraints. 
\end{definition} 
\begin{definition}[Total Loss for Utility Model]
\label{total_loss}
    \begin{align}
        \mathcal{L}_{\text{Total}} =\mathcal{L}_{\text{Rank}_{12}} + \lambda_{\text{OT}} \cdot \mathcal{L}_{\text{OT}}
    \end{align}
where $\lambda_{\text{OT}}$ is a hyperparameter. 
\end{definition}

\textbf{c) How do we collect utility samples iteratively?}
The very first question encountered during pretraining is how to generate utility samples. \citet{ilyas2022datamodels} construct training subsets by random sampling a fixed-length subset. 
One caveat in our setting is the growing length of labeled set as the progression of the active learner. To enable the model to adapt to growing length of utility samples, one needs to incorporate \textit{diversity} in the size of $\utilitysample$. 
One natural choice is to perform rejection sampling from the \textit{powerset} of $\LabeledSet_{0}$, i.e., $\utilitysample \sim 2^{\LabeledSet_{0}}$. Instead of fixing the sampling proportion, 
we propose to fix the number of utility samples collected from $\LabeledSet_{0}$ per iteration during pretraining as $n$.

\textbf{d) How do we update the set-based NN during pretraining?}
As mentioned in Section \ref{surrogatemodel}, the length of labeled utility samples grows and random split for training and validation set may fail to capture the notion of generalizability in neural batch active learning. The goal of utility model is to \textit{generalize} to longer length of utility samples and learn a general mapping from utility sample to validation accuracy. Inspired by bilevel training work \citep{franceschi2018bilevel, grazzi2020iteration, borsos2021semi}, we employ a bilevel framework to separate the utiltiy samples by \textit{length}. In practice, we separate out the validation set and training set by $50\%$ and $50\%$ for simplicity.  We retrain the set-based NN per iteration with the accumulation of utility samples per iteration. We defer the complete discussion of bi-level training to Section~\ref{bilevel_opt}.

\textbf{e) How do we acquire data in the acquisition stage?} 
In the context of utility maximization, perhaps the simplest candidate is to select the instance with largest predicted utility. Popular approaches rely on sequentially picking one data point per round \citep{houlsby2011bayesian, gal2017deep} though the addition of single data point cause minimal change to validation accuracy while increasing the cost of model retraining. \citet{alieva2020learning} suggest that for many sequential decision making problems, a data-driven greedy heuristics for sequentially selecting actions that is regularized to have diminishing returns exhibits superior performance 
without invoking expensive evaluation oracles. Recall that one shall interpret $\hat{u}$ as a score-based acquisition function and leverage it for sequential decision making, i.e. to greedily select the utility sample with highest predicted value by comparing the utility between pairs of utility samples. Inspired by \citet{citovsky2021batch}, we employ Margin Sampling \citep{roth2006margin} as a filter for unlabeled instance i.e., select $M$ unlabeled instances with lowest margin scores, per iteration in the acquisition stage (See Algorithm~\ref{alg:Greedy-Margin}). We propose to randomly split $\Unlabeled_{0}$ into batches of size $b$, concatenate each batch to the current labeled pool, and then use the concatenated batch as input to $\hat{u}$ for 
utility prediction. We perform sequential batch selection within the acquisition stage and select the unlabeled batch with the largest predicted score.
\subsection{The RAMBO Algorithm}
The essence of our two-stage utility model aligns with Shakespeare’s famous line from The Tempest, “What’s past is prologue.” 
Our overarching motivation is to train an acquisition function on past utility samples 
that generalizes well to utility samples of longer history. 
We first initialize the surrogate utility model, 
by training an initial parametrized utility model $\hat{u}$ from offline datasets, providing an initial estimate of the 
\textit{feature extractor} $\phi_{0}$. 
This initial feature extractor $\phi_{0}(\cdot)$ can serve as a warm start for non-adaptive batch selection in the acquisition stage. 
We emphasize the need for this \textit{initialization} step as \algname is developed for single-round selection. 
\begin{algorithm}[t]
\caption{\algname}
\begin{algorithmic}[1]
\State {\bf Input}: 
$B$, 
$\Unlabeled_{0}$, 
$\LabeledSet_{0}$ 
$\mathcal{X}$, 
$b$, 
$M$, 
$n$, 
$\LabeledSet_{val}$.
\State {\bf Output}: $\LabeledSet_{1}$
\State Initialize $(\hat{u}_{0}, \phi_{0})$ from offline dataset 
\State Randomly divide $\LabeledSet_{0}$ into $S_{0}$ with size $k_{1}$ and $\{s_{1}, s_{2} ... s_{\tau_{1}}\}$ with each size  
$b$ and set $U_{0} = \Unlabeled_{0}$
\State $\tau_{1} = \frac{k-k_{1}}{b}$ and $\tau_{2} = \frac{B}{b}$
\State Train $f$ on $S_{0}$ and get accuracy on $\LabeledSet_{val}$ as $acc_{0}$
\State $\mathcal{D}_{0} \leftarrow \{\}$
\For{$i = 0: \tau_{1}$} 
\Comment{\textbf{Pretraining}}
    \State $S_{i+1} \leftarrow S_{i} \cup \{ s_{i+1} \}$
    \State Train $f$ on $S_{i+1}$
    \State Obtain accuracy on $\LabeledSet_{val}$ as $acc_{i+1}$
    \State $D_{i+1} \leftarrow$ Utility-Samples-Augmentation($S_{i},$ \\
$S_{i+1}, n, acc_{i}, acc_{i+1}, D_{i}$) 
    \State Train $\hat{u}_{i}$ from $D_{i+1}$ 
    \Comment{\textbf{Bilevel Optimization}}
\EndFor
\For {$j = 0: \tau_{2}$} \Comment{\textbf{Acquisition}}
    \State $S_{j+1}, U_{j+1} \leftarrow \text{Greedy-Margin}(\hat{u}_{\tau_{1}}, j, b, S_{j}, M, U_{j})$
\EndFor
\State $\LabeledSet_{1}, \Unlabeled_{1} = S_{\tau_{2}}, U_{\tau_{2}}$ 
\end{algorithmic}
\end{algorithm}
\begin{algorithm}[t]
    \caption{Greedy-Margin}
\label{alg:Greedy-Margin}
    \begin{algorithmic}[1]
        \State {\bf Input}: $\hat{u}$, $j$, $b$, $S_{j}$, $M$, $U_{j}$.
        \State {\bf Output}: $S_{j+1}$, $U_{j + 1}$
            \State $R \rightarrow$ a subset obtained by smallest margin scores $M$ examples from $U_{j} \setminus S_{j}$
            \State Randomly divide $R$ into $\{ \lfloor \frac{R}{b} \rfloor \}$ batches of subsets $\{ (x_{i})_{i=1}^{b} \}$.
            \State $b_{\max} \leftarrow \argmax_{ \{ (x_{i})_{i=1}^{b} \} \in \{ \lfloor \frac{R}{b} \rfloor \}} \hat{u}(S_{j} \cup (x_{i})_{i = 1}^{b})$
            \State $S_{j + 1} \leftarrow S_{j} \cup \{ b_{\max}\}$
            \State $U_{j+1} \leftarrow U_{j} \setminus \{ b_{\max} \}$
    \end{algorithmic}
\end{algorithm}
\begin{algorithm}[t]
    \caption{Utility-Samples-Augmentation}
    \label{alg:interpolate}
    \begin{algorithmic}[1]
        \State {\bf Input}: $S_i$, $S_{i+1}$, $n$, $acc_{i}$, $acc_{i+1}$, $D_{i}$.
        \State {\bf Output}: $D_{i}$
        \For{$i \in \text{range}(n)$}
        \State Sample random a pair of $(\utilitysample_{1}, \utilitysample_{2})$ from $S_{i}$ with equal size
        \State Compute distance between $\phi(\utilitysample_{1})$ and $\phi(S_{i})$ as $d_{1,i}$ and distance between $\phi(\utilitysample_{1})$ and $\phi(S_{i+1})$ as $d_{1, i+1}$. Same Rule applies to $\utilitysample_{2}$ to obtain $d_{2,i}$ and $d_{2,i+1}$.
        \State Calculate $u_{1}$, $u_{2}$ for $\utilitysample_{1}$ and $\utilitysample_{2}$ by Equation~\ref{interpolation}
        \State $D_{i} \leftarrow D_{i} \cup \{ (\utilitysample_{1}, u_{1}), (\utilitysample_{2}, u_{2}) \}$
        \EndFor
    \end{algorithmic}
\end{algorithm}
\subsubsection{Bi-Level Optimization}
\label{bilevel_opt}
To align with growing labeled pool of active learning setting, a core requirement of our utility model is the capability to \textit{generalize to longer and unseen data} by drawing on prior utility samples.
A line of research \citep{rajeswaran2019meta, liu2019self} suggests that meta-learning shall lead to fast adaptation and generalization to new tasks. One formulation of meta-learning is bi-level optimization \citep{maclaurin2015gradient} where the inner objective represents adaptation to a given task and the outer problem is the meta-training objective.
Motivated by \citet{franceschi2018bilevel}, we formulate utility model training as bilevel optimization, combining gradient-based hyperparameter optimization and meta-learning in which the outer optimization problem is solved subject to the optimality of an inner optimization problem. To improve the utility model's generalization capability on samples with varied lengths, we divide the utility samples  $(\utilitysample, u(\utilitysample))$ at iteration $i$ to training $D_{tr}$ and validation set $D_{val}$ by length, where $D_{tr}$ corresponds to utility samples with length smaller than the median and vice versa, and treat them as input dataset for \textit{inner objective} $L$ and \textit{outer objective} $E$. Formally, we consider the bilevel optimization framework as 
\begin{align}
\min_{\lambda}~ E(w(\lambda), \lambda) 
\text{~~s.t. ~} w(\lambda) = \argmin_{\hat{w} \in \mathbb{R}^{d}} 
    \mathcal{L}(\hat{w}) \nonumber
\end{align}
where $\lambda$ is a hyperparameter 
, $E$ and $\mathcal{L}$ are continuously differentiable functions, the outer objective
\begin{align}
    E(w(\lambda), \lambda) := \sum\limits_{ \{(S_{1}', u(S_{1}')), (S_{2}', u(S_{2}'))\} \in D_{val}} \mathcal{L}_{\text{Total}}(\hat{w}) 
    \nonumber
\end{align}
and the inner objective as
\begin{align}
     \begin{split}
     \mathcal{L}(\hat{w}) = \sum\limits_{\{(S_{1}', u(S_{1}')), (S_{2}', u(S_{2}'))\} \in D_{tr}} \mathcal{L}_{\text{Total}}(\hat{w}) + \Omega_{\lambda}(\hat{w}) \nonumber
\end{split}
\end{align}
where $D_{\text{tr}} = \{(\utilitysample_{1}, u(\utilitysample_{1})), (\utilitysample_{2}, u(\utilitysample_{2})) \}_{i=1}^{n}$ is a set of pair of utility samples attributed to training set and $\mathcal{L}_{\text{Total}}(\cdot)$ is the BCE loss induced by the supervised algorithm and $\Omega_{\lambda}$ is a regularizer parametrized by $\lambda$. The outer objective is the proxy of generalization error of $\hat{u}(\cdot)$, given by the average loss of $D_{\text{val}}$.

The inner optimization is aimed at \emph{utility model optimization}, i.e., finding the best model parameters that minimize the total loss on smaller length training samples $D_\text{tr}$. Conversely, the outer optimization targets to \emph{generalize the model to longer-length utility samples} $D_\text{val}$, which seeks the optimal regularizer parameterized by $\lambda$. With this bilevel formulation, RAMBO shows better and more stable performance when performing unlabeled data selection on CIFAR10 with labeling budge 5000 (as suggested by Table~\ref{BilevelTraining1}).
Table~\ref{BilevelTraining1} shows average performance of models with bilevel training used in optimization, mostly outperforms the rest of counterparts without bilevel training, illustrating the enhanced generalizability across various model architectures and training algorithms.
\subsubsection{Interpolation-Based Utility Samples}\label{interpolation based surrogates}
Yet, the scarcity of utility samples poses challenges to the efficacy of our utility model training. To alleviate the need for groundtruth utility samples, we leverage the consistency regularization techniques from semi-supervised learning to augment artificial $(\utilitysample, u(\utilitysample))$. 
Inspired by \citet{parvaneh2022active}, the latent space of the classifier's feature extractor shall contain valuable representations that can be interpolated within labeled instances. The empirical success suggests a change in perspective---rather than twisting the classifier, we leverage the shared representations in $\hat{u}$ throughout the progress of optimization. In particular, we adopt the \textit{interpolation consistency regularization} strategy \citep{verma2022interpolation} (Definition~\ref{def:interp}). The pseudo code for utility samples augmentation is outline \ref{alg:interpolate}.  
\begin{definition}[Utility Value Interpolation]
Denote the validation accuracy at iteration $i$ as $acc_{i}$. 
For a given utility sample $\utilitysample_{1}$, let $d_{1,i}$ be its distance with the previous labeled pool $S_i$ and $d_{1,i+1}$ the distance with the current labeled pool $S_{i+1}$. The augmented utility value $u_{1}$ for $\utilitysample_{1}$ yields as
\begin{align}
\label{interpolation}
    u_{1} &= \alpha \cdot u_{i} + (1 - \alpha) \cdot u_{i+1}
    \vspace{-2mm}
\end{align}
with $\alpha := \frac{d_{1,i+1}}{d_{1,i+1} + d_{1,i}}.$\label{def:interp}
\end{definition}
\label{Method}
\section{Experimental Results}\label{Experiments}

\subsection{Experimental Setup}

\label{experimental_setup}
Here, we evaluate the performance of \algname against several state-of-the-art baselines on four image datasets MNIST \citep{lecun1998gradient}, FashionMNIST \citep{xiao2017fashion}, CIFAR10 \citep{krizhevsky2009learning}, SVHN\citep{netzer2011reading}. To ensure a comprehensive comparison among all algorithms, we evaluate them across various acquisition stage budget $B$ as \{500, 700, 900, 1000\} for MNIST and FashionMNIST with $k = 200$, \{5000, 7000, 9000, 10000\} for CIFAR10 and SVHN with $k = 2500$. For main results, we focus on the accuracy of validation set as the key performance metric. We fix the validation size to be 1000 across all datasets. Lastly, we run each experiment ten times and report average and standard error across all experiments.
Depending on the type of dataset, we consider different network architectures for classifiers. We consider two classifier structures: one is \citet{beck2021effective}'s neural network structure, a model similar to LeNet \citep{lecun1998gradient} for MNIST and FashionMNIST 
and the other is ResNet-18 \citep{he2016deep} for CIFAR10 and SVHN. We provide details of utility model architecture to the Appendix. 

We fit all classifiers using cross-entropy loss with optimizer Adam until training accuracy exceeds $99\%$ with maximum $100$ epochs and learning rate $0.001$. No learning rate schedulers and data augmentations are used. Baselines use implementations from open-source AL toolkit DISTIL \citet{decileteam_2023_distil}. All models are trained in PyTorch \citep{paszke2017automatic}.

\subsection{Baselines}
For all experiments, we consider a set of baselines that consists of a classical Margin Sampling algorithm, as well as two recent active learning, BADGE and CoreSet, one learning-based algorithm GLISTER and random selection Random.

\textbf{Margin Sampling}: Selects $B$ examples from $\Unlabeled_{0}$ with smallest difference between the first and second most probable classes predicted by $f$ \citep{roth2006margin}. 

\textbf{BADGE}: A hyperparameter-free approach that trades off between diversity and uncertainty using k-means$++$ in hallucinated gradient space \citep{ash2019deep}.

\textbf{CoreSet}: A diversity based approach using greedy approximation to the k-center problem on representations from the current classifier's penultimate layer \citep{sener2017active}.   

\textbf{GLISTER}: A learning-based approach selecting $B$ instances from $\LabeledSet_{0}$ that would maximize the log-likelihood on held-out validation set $\LabeledSet_{val}$ by converting it as a mixed discrete continuous bilevel optimization. We adopt the GLISTER-ONLINE version as an approximation for the inner optimization problem by taking a single gradient step update \citep{killamsetty2021glister}. 

\textbf{Random}: Selects $B$ samples uniformly at random. 

\begin{figure*}[t!]
\centering
    \begin{subfigure}{.31\textwidth}
        \centering
\includegraphics[clip,trim=0cm 0cm 0cm 0cm,width=\textwidth]{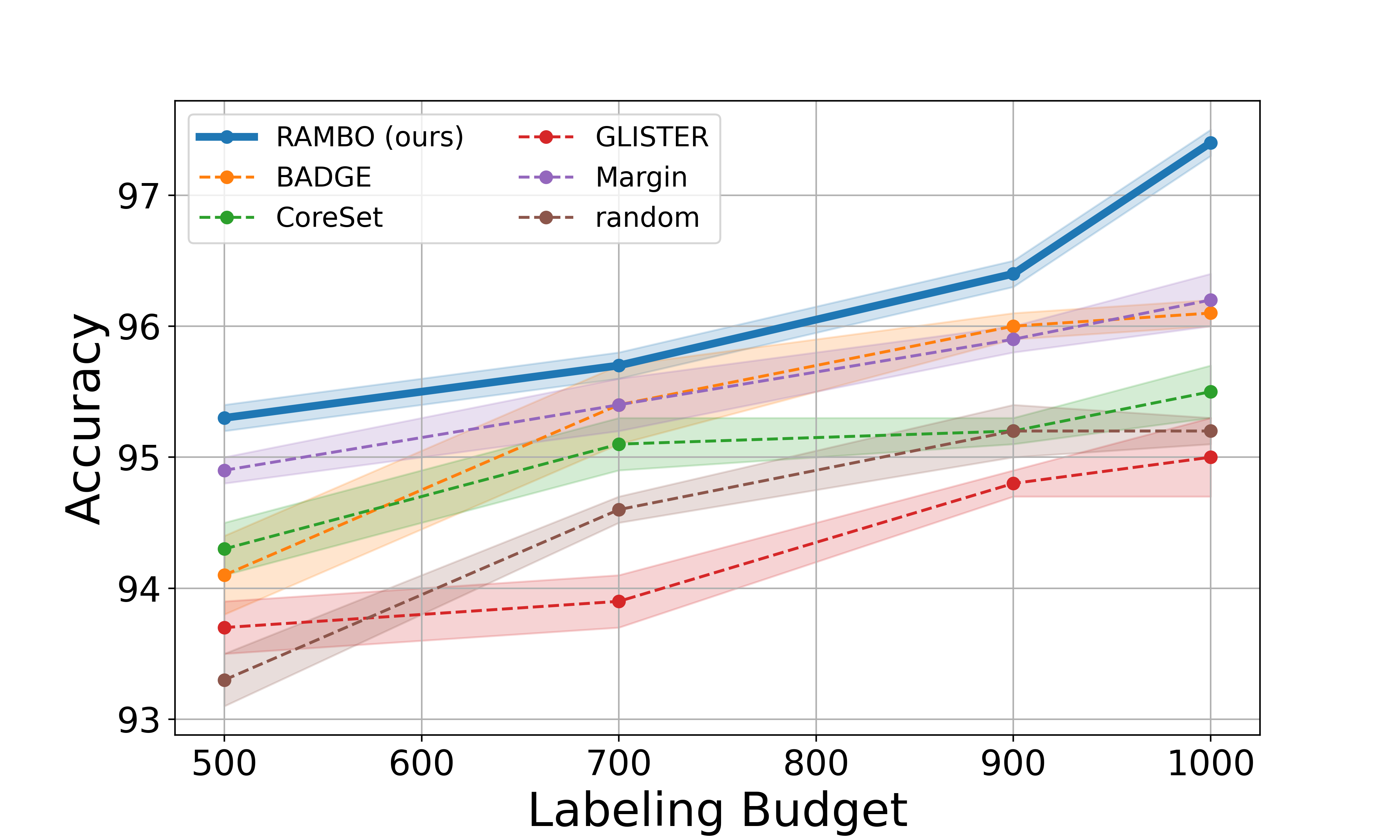}
    \caption{\footnotesize MNIST}
    \vspace{-2mm}
    \label{Pretraining Budget Variation}
     \end{subfigure}
    \begin{subfigure}{.31\textwidth}
        \centering
        \includegraphics[clip,trim=0cm 0cm 0cm 0cm,width=\textwidth]{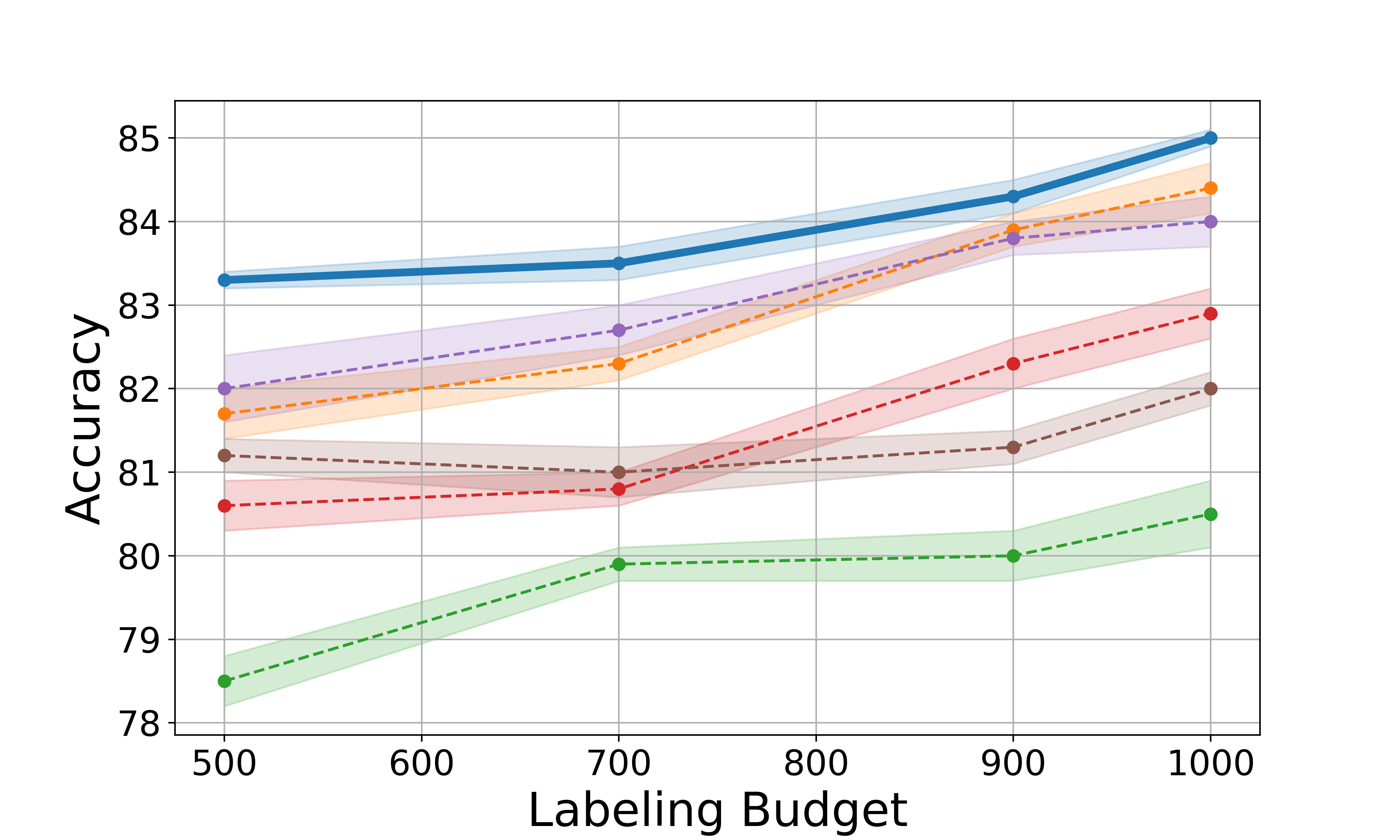}
        \caption{FashionMNIST}\label{}
        \vspace{-2mm}
    \end{subfigure}
    \begin{subfigure}{.31\textwidth}
        \centering
        \includegraphics[clip,trim=0cm 0cm 0cm 0cm,width=\textwidth]{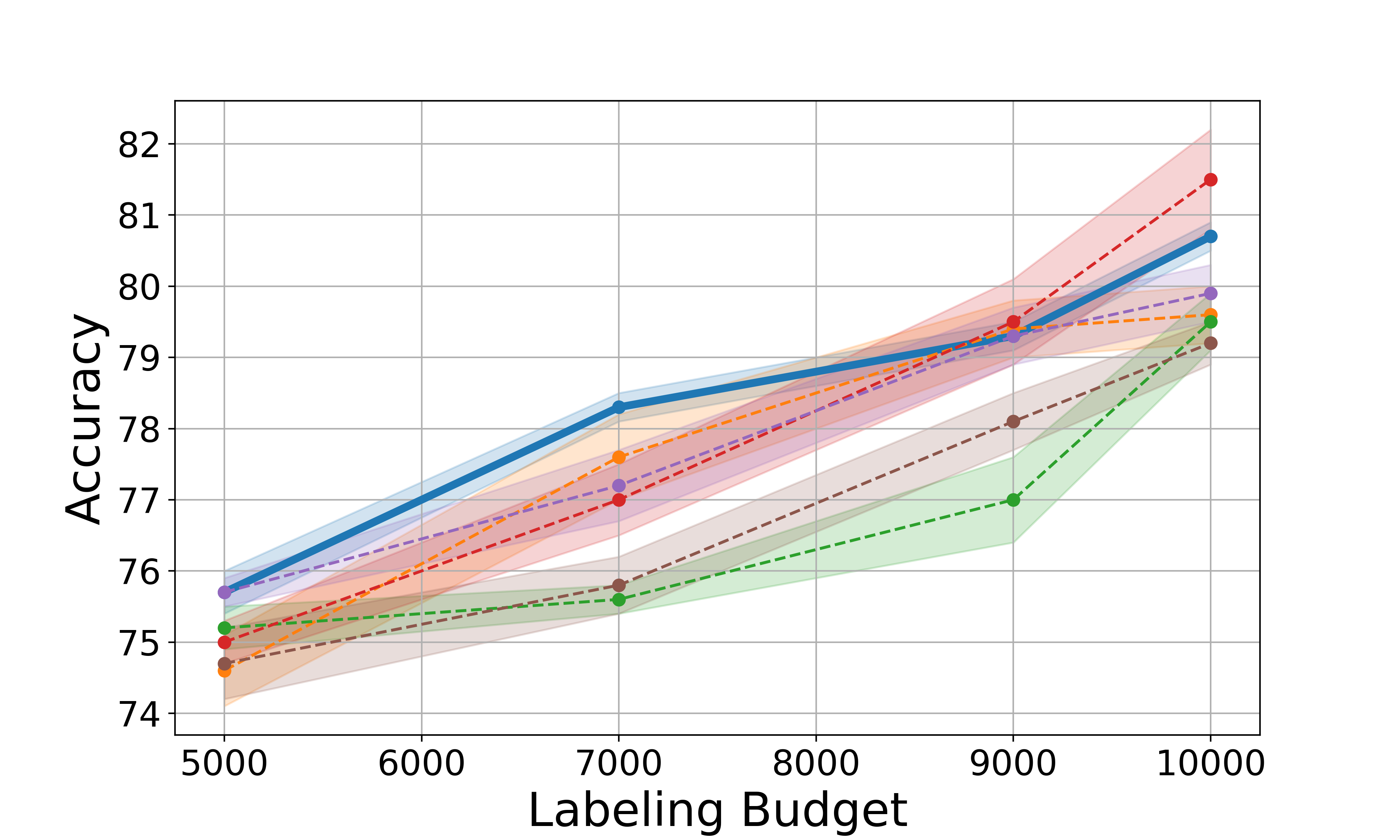}
        \caption{CIFAR10}\label{}
        \vspace{-2mm}
    \end{subfigure}
    \\
    \begin{subfigure}{.31\textwidth}
        \centering
        \includegraphics[clip,trim=0cm 0cm 0cm 0cm,width=\textwidth]{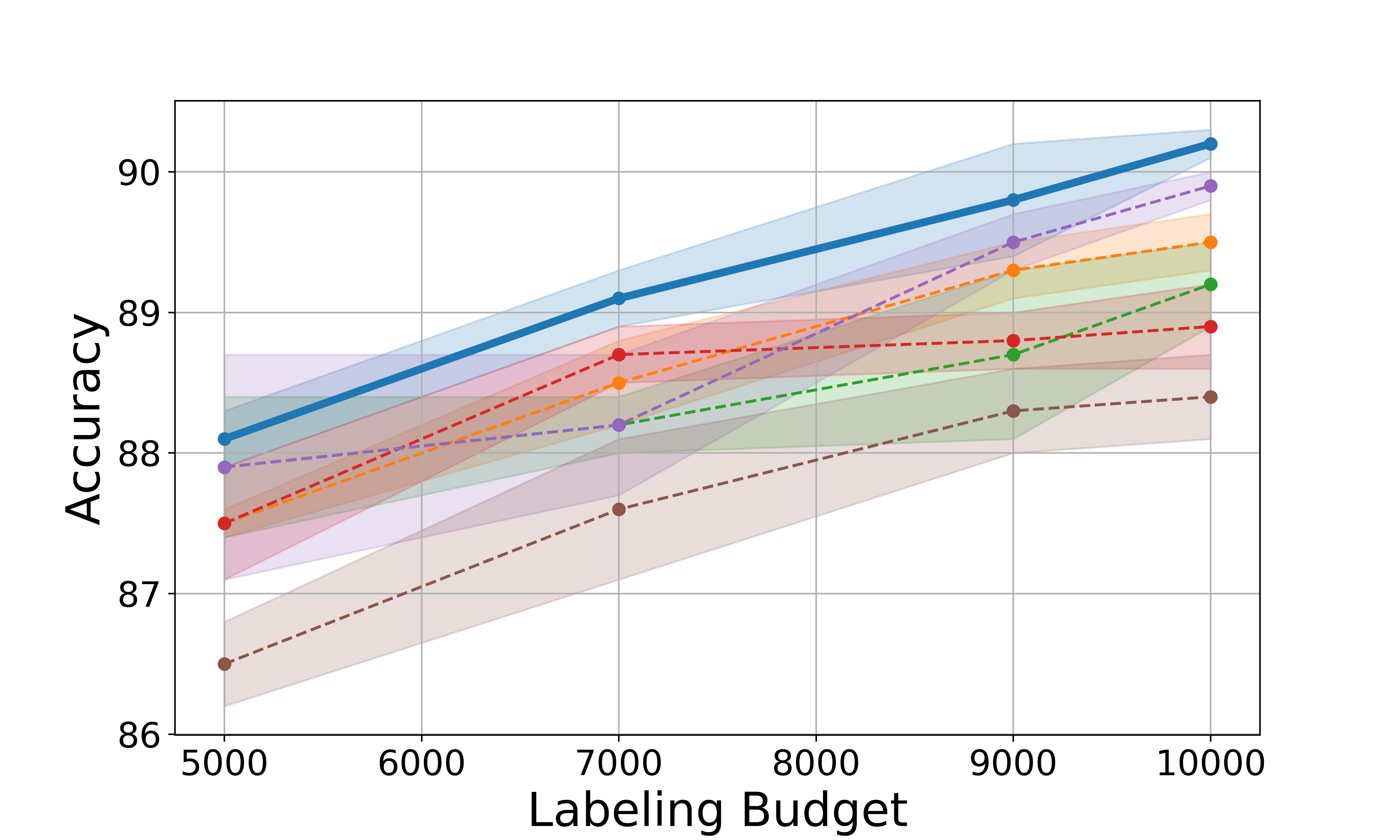}
        \caption{SVHN}\label{}
        \vspace{-1mm}
    \end{subfigure}
    \begin{subfigure}{.31\textwidth}
    \includegraphics[clip,trim=0cm 0cm 0cm 0cm,width=\textwidth]{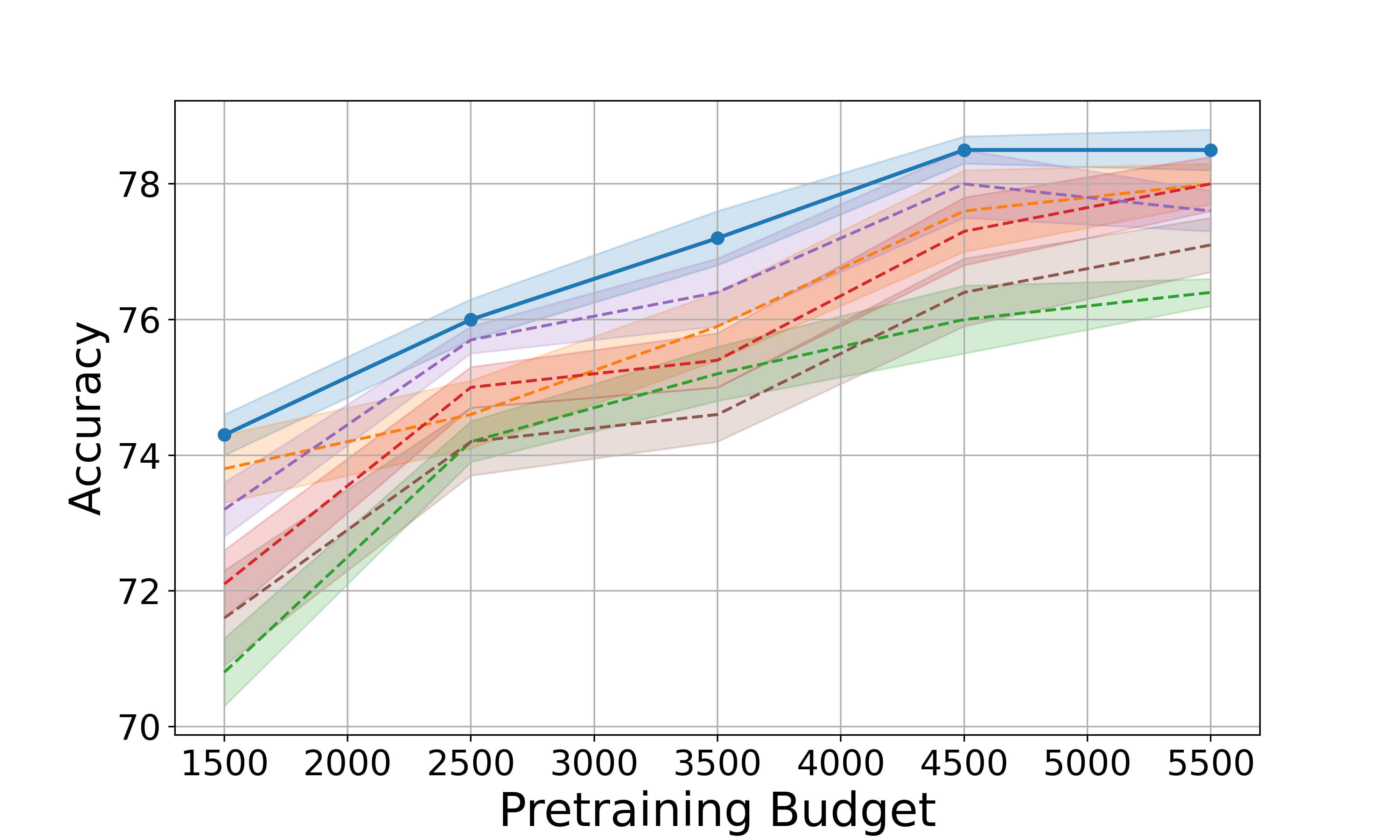}
        \caption{ Ablation on $k$
        }
        \vspace{-1mm}
    \label{CIFAR10_Pretraining_Budget_Variation}
    \end{subfigure}
    \begin{subfigure}{.31\textwidth}
\includegraphics[clip,trim=0cm 0cm 0cm 0cm,width=.9\textwidth]{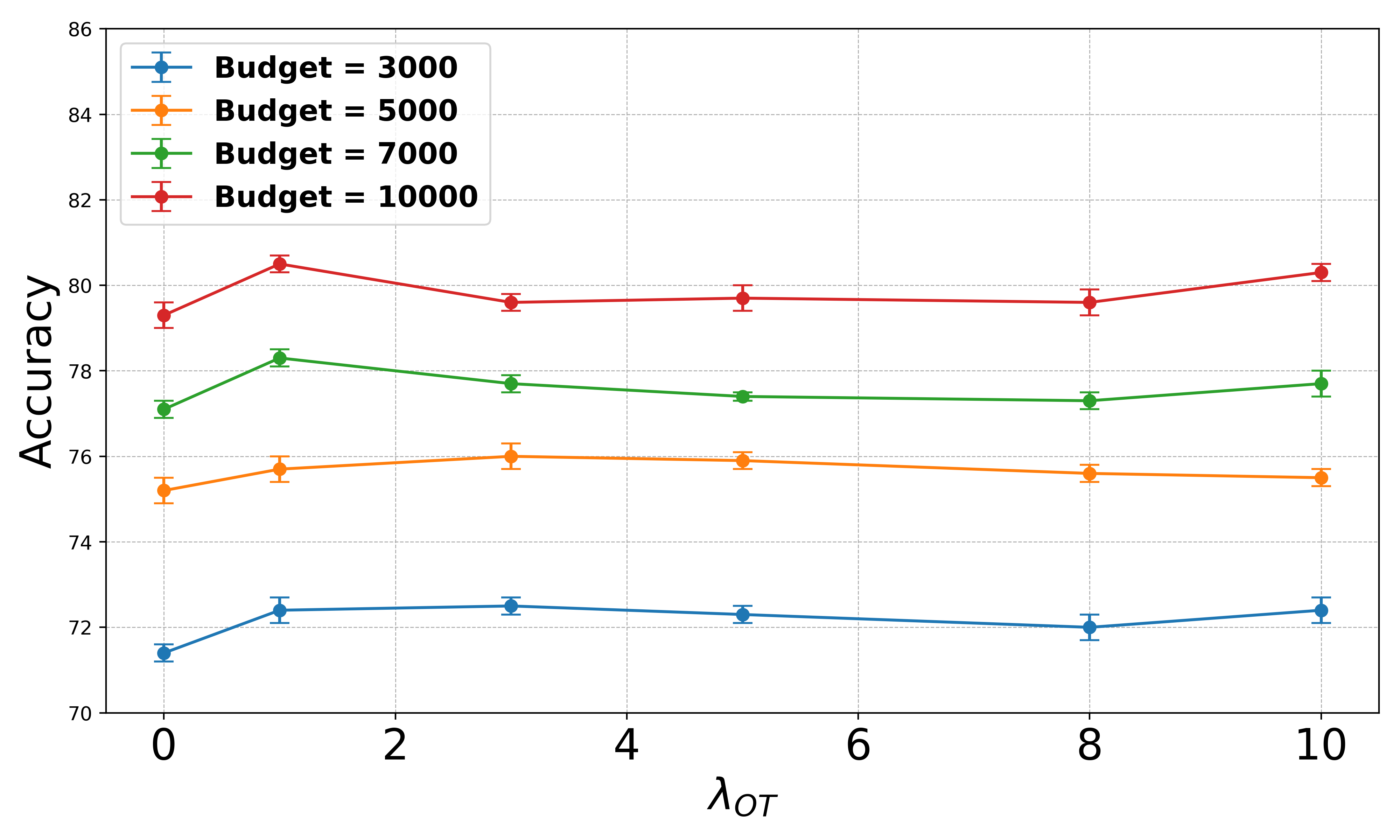}
    \caption{Ablation on $\lambda_{\text{OT}}$
    }
    \vspace{-3mm}
    \label{lambda_OT hyperparameter}
    \end{subfigure}
    \caption{Experimental results. \textbf{(a-d)} Active learning validation performance. \textbf{(e)} Active Learning validation performance with the acquisition stage budget $B=5000$ for CIFAR10 across various choices of pretraining budget $k$. \textbf{(f)} Different choices of $\lambda_{\text{OT}}$ for pretraining set size $k=2500$ on CIFAR10. Results are given in \%.} 
    \label{Accuracy Validation Performance}
\end{figure*}

\subsection{Results}
In Figure \ref{Accuracy Validation Performance}, \algname outperforms most of baselines across multiple architectures and various labeling budget for acquisition stage. For easy datasets like FashionMNIST and MNIST, \algname shall learn a good shared representation for effective utility value interpolation and can easily beats all the baselines oblivious to different labeling budgets which suggests \algname is a good choice regardless of labeling budget. As BADGE and CoreSet operate on the penultimate layer with limited budget, both algorithms fail to perform well as their learned representations might not be accurate. However, \algname performs interpolation techniques to augment utility samples within limited labeled pool and generalize to predictions of longer history of labeled data, leading to learning-based acquisition function amenable to growing labeled pool.

Even for hard datasets, for CIFAR10 and SVHN, when the model fails to have a good architecture priors due to limited pool, BADGE and CoreSet cannot learn meaningful representations. Occationally, CoreSet might not outperform passive learning, for instance, labeling Budget is 7000 for CIFAR10.

\subsection{Ablation Study}
\label{ablation}

We perform an ablation study on the size of Pretraining set, the design choices of each submodule as bilevel training, OT distance and RankNet as well as hyperparameter for OT Distance Loss (Definition~\ref{total_loss}). We use CIFAR10 as an example dataset, and defer our results on the remaining datasets to the Appendix.

\looseness -1 \paragraph{Size of Pretraining Budget}
Naturally, we want to examine the effect of size of pretraining set for determining how the scale of initial labeled pool impacts overall single round selection performance. Figure~\ref{CIFAR10_Pretraining_Budget_Variation} shows across different seed set size for pretraining stage, \algname outperforms all other baselines.

\paragraph{Bi-level training, OT Distance and RankNet}
\label{Three Design Choices}
Next, we shift to study the intertwined effects of three design choices. 
Table~\ref{BilevelTraining1} shows the combined efficacy of bilevel training, OT distance, and RankNet, offering insights into the synergy of these three foundational modules. The cross mark for RankNet means regression based acquisition function and the loss is designed as MSE between predicted utility vs. true utility value. One thing to note is that if the performance of regression based acquisition function without bi-level training and OT distance is similar to random, which corroborates our intuition about ranking instead of regressing validation accuracy on labeled samples.
\begin{table}[h!]
\centering
\small
\caption{Ablation study on three submodules with pretraining set $k=3500$ and acquisition budget $B=5000$. The last row 
corresponds to the random baseline.}
\vspace{-2mm}
\label{BilevelTraining1}
\scalebox{0.85}{
\begin{tabular}{lccc}
    \toprule
    Bilevel & Optimal Transport & RankNet & Accuracy\\
    \midrule
     $\checkmark$ & $\checkmark$ & $\checkmark$ & $\mathbf{77.3 \pm 0.2}$ \\
    $\checkmark$ & $\checkmark$ & $\times$ & $76.1 \pm 0.3$\\
    $\checkmark$ & $\times$ & $\checkmark$ & $76.2 \pm 0.4$\\
    $\checkmark$ & $\times$ & $\times$ & $70.5 \pm 0.3$ \\
    $\times$ & $\checkmark$ & $\checkmark$ & $75.5 \pm 0.3$\\
     $\times$ & $\checkmark$ & $\times$ & $75.5 \pm 0.3$ \\
    $\times$ & $\times$ & $\checkmark$ & $76.0 \pm 0.8$ \\
    $\times$ & $\times $ & $\times$ & $74.6 \pm 0.7$ \\
    - & - & - & $74.7 \pm 0.3$ \\
    \bottomrule
\end{tabular}}
\vspace{-3mm}
\end{table}




\vspace{-2mm}
\paragraph{Hyperparameter Tuning for OT distance}
By definition, $\mathcal{L}_{\text{Total}} = \mathcal{L}_{\text{Rank}_{12}} + \lambda_{\text{OT}} \cdot \mathcal{L}_{\text{OT}}$ (Definition \ref{total_loss}). One can change the scale of $\lambda_{\text{OT}}$ for utility model training in pretraining. We study the effect of hyperparameter $\lambda_{\text{OT}}$ in final model performance on validation set. We highlight the importance of incorporating OT distance into the loss structure which makes $\hat{u}$ insensitive to the scale of $\lambda_{\text{OT}}$. When $\lambda_{\text{OT}} > 0$, the overall validation accuracy is larger than $\lambda_{\text{OT}} = 0$. The choice of $\lambda_{\text{OT}}$ is specific to dataset and batch setting and we present one setting of $\lambda_{\text{OT}}$ with varied Labeling Budget for acquisition stage in Figure \ref{lambda_OT hyperparameter}.
\section{Conclusion}
We have demonstrated existing state-of-the-art methods shall be suboptimal in single round selection. We show that under certain budget for pretraining, \algname shall achieve better generalization performance compared to other active learning algorithms, and that most of validation accuracy improvement is realized by our two-stage algorithm. Finally, we illustrate how behaviors of all algorithms change with variation of pretraining and single round acquisition budget across multiple datasets and architectures.
One potential direction for future work could be to determine an optimal budget allocation for both the pretraining and acquisition stages, as well as the extension of \algname to the few-rounds setting. 

\section*{Acknowledgements}
We are thankful to Feiyang Kang and Jiachen T. Wang for providing useful discussion and helpful feedback on the paper. YC and ZD acknowledge support from the National Science Foundation under Grant No. NSF IIS-2313131 and NSF FMRG-2037026. RJ and the ReDS lab acknowledge support through grants from the Amazon-Virginia Tech Initiative for Efficient and Robust Machine Learning, the National Science Foundation under Grant No. IIS-2312794, NSF IIS-2313130, NSF OAC-2239622, and the Commonwealth Cyber Initiative.

\bibliographystyle{plainnat}
\bibliography{reference.bib}

\begin{thebibliography}{65}
\providecommand{\natexlab}[1]{#1}
\providecommand{\url}[1]{\texttt{#1}}
\expandafter\ifx\csname urlstyle\endcsname\relax
  \providecommand{\doi}[1]{doi: #1}\else
  \providecommand{\doi}{doi: \begingroup \urlstyle{rm}\Url}\fi

\bibitem[Adaimi and Thomaz(2019)]{adaimi2019leveraging}
Rebecca Adaimi and Edison Thomaz.
\newblock Leveraging active learning and conditional mutual information to
  minimize data annotation in human activity recognition.
\newblock \emph{Proceedings of the ACM on Interactive, Mobile, Wearable and
  Ubiquitous Technologies}, 3\penalty0 (3):\penalty0 1--23, 2019.

\bibitem[Alieva et~al.(2020)Alieva, Aceves, Song, Mayo, Yue, and
  Chen]{alieva2020learning}
Ayya Alieva, Aiden Aceves, Jialin Song, Stephen Mayo, Yisong Yue, and Yuxin
  Chen.
\newblock Learning to make decisions via submodular regularization.
\newblock In \emph{International Conference on Learning Representations}, 2020.

\bibitem[Altschuler et~al.(2017)Altschuler, Niles-Weed, and
  Rigollet]{altschuler2017near}
Jason Altschuler, Jonathan Niles-Weed, and Philippe Rigollet.
\newblock Near-linear time approximation algorithms for optimal transport via
  sinkhorn iteration.
\newblock \emph{Advances in neural information processing systems}, 30, 2017.

\bibitem[Alvarez-Melis and Fusi(2020)]{alvarez2020geometric}
David Alvarez-Melis and Nicolo Fusi.
\newblock Geometric dataset distances via optimal transport.
\newblock \emph{Advances in Neural Information Processing Systems},
  33:\penalty0 21428--21439, 2020.

\bibitem[Ash et~al.(2019)Ash, Zhang, Krishnamurthy, Langford, and
  Agarwal]{ash2019deep}
Jordan~T Ash, Chicheng Zhang, Akshay Krishnamurthy, John Langford, and Alekh
  Agarwal.
\newblock Deep batch active learning by diverse, uncertain gradient lower
  bounds.
\newblock \emph{arXiv preprint arXiv:1906.03671}, 2019.

\bibitem[Bachman et~al.(2017)Bachman, Sordoni, and
  Trischler]{learnalgoforactivelearning}
Philip Bachman, Alessandro Sordoni, and Adam Trischler.
\newblock Learning algorithms for active learning.
\newblock \emph{CoRR}, abs/1708.00088, 2017.
\newblock URL \url{http://arxiv.org/abs/1708.00088}.

\bibitem[Beck et~al.(2021)Beck, Sivasubramanian, Dani, Ramakrishnan, and
  Iyer]{beck2021effective}
Nathan Beck, Durga Sivasubramanian, Apurva Dani, Ganesh Ramakrishnan, and
  Rishabh Iyer.
\newblock Effective evaluation of deep active learning on image classification
  tasks.
\newblock \emph{arXiv preprint arXiv:2106.15324}, 2021.

\bibitem[Borsos et~al.(2021)Borsos, Tagliasacchi, and Krause]{borsos2021semi}
Zal{\'a}n Borsos, Marco Tagliasacchi, and Andreas Krause.
\newblock Semi-supervised batch active learning via bilevel optimization.
\newblock In \emph{ICASSP 2021-2021 IEEE International Conference on Acoustics,
  Speech and Signal Processing (ICASSP)}, pages 3495--3499. IEEE, 2021.

\bibitem[Botu and Ramprasad(2015)]{botu2015adaptive}
Venkatesh Botu and Rampi Ramprasad.
\newblock Adaptive machine learning framework to accelerate ab initio molecular
  dynamics.
\newblock \emph{International Journal of Quantum Chemistry}, 115\penalty0
  (16):\penalty0 1074--1083, 2015.

\bibitem[Burges et~al.(2005)Burges, Shaked, Renshaw, Lazier, Deeds, Hamilton,
  and Hullender]{burges2005learning}
Chris Burges, Tal Shaked, Erin Renshaw, Ari Lazier, Matt Deeds, Nicole
  Hamilton, and Greg Hullender.
\newblock Learning to rank using gradient descent.
\newblock In \emph{Proceedings of the 22nd international conference on Machine
  learning}, pages 89--96, 2005.

\bibitem[Chen et~al.(2015{\natexlab{a}})Chen, Hassani, Karbasi, and
  Krause]{chen2015sequential}
Yuxin Chen, S~Hamed Hassani, Amin Karbasi, and Andreas Krause.
\newblock Sequential information maximization: When is greedy near-optimal?
\newblock In \emph{Conference on Learning Theory}, pages 338--363. PMLR,
  2015{\natexlab{a}}.

\bibitem[Chen et~al.(2015{\natexlab{b}})Chen, Javdani, Karbasi, Bagnell,
  Srinivasa, and Krause]{chen2015submodular}
Yuxin Chen, Shervin Javdani, Amin Karbasi, J~Bagnell, Siddhartha Srinivasa, and
  Andreas Krause.
\newblock Submodular surrogates for value of information.
\newblock In \emph{Proceedings of the AAAI Conference on Artificial
  Intelligence}, volume~29, 2015{\natexlab{b}}.

\bibitem[Citovsky et~al.(2021)Citovsky, DeSalvo, Gentile, Karydas, Rajagopalan,
  Rostamizadeh, and Kumar]{citovsky2021batch}
Gui Citovsky, Giulia DeSalvo, Claudio Gentile, Lazaros Karydas, Anand
  Rajagopalan, Afshin Rostamizadeh, and Sanjiv Kumar.
\newblock Batch active learning at scale.
\newblock \emph{Advances in Neural Information Processing Systems},
  34:\penalty0 11933--11944, 2021.

\bibitem[Coleman et~al.(2019)Coleman, Yeh, Mussmann, Mirzasoleiman, Bailis,
  Liang, Leskovec, and Zaharia]{coleman2019selection}
Cody Coleman, Christopher Yeh, Stephen Mussmann, Baharan Mirzasoleiman, Peter
  Bailis, Percy Liang, Jure Leskovec, and Matei Zaharia.
\newblock Selection via proxy: Efficient data selection for deep learning.
\newblock \emph{arXiv preprint arXiv:1906.11829}, 2019.

\bibitem[D'Amour et~al.(2022)D'Amour, Heller, Moldovan, Adlam, Alipanahi,
  Beutel, Chen, Deaton, Eisenstein, Hoffman, et~al.]{d2022underspecification}
Alexander D'Amour, Katherine Heller, Dan Moldovan, Ben Adlam, Babak Alipanahi,
  Alex Beutel, Christina Chen, Jonathan Deaton, Jacob Eisenstein, Matthew~D
  Hoffman, et~al.
\newblock Underspecification presents challenges for credibility in modern
  machine learning.
\newblock \emph{The Journal of Machine Learning Research}, 23\penalty0
  (1):\penalty0 10237--10297, 2022.

\bibitem[Fang et~al.(2017)Fang, Li, and Cohn]{learntoactivelearning}
Meng Fang, Yuan Li, and Trevor Cohn.
\newblock Learning how to active learn: {A} deep reinforcement learning
  approach.
\newblock \emph{CoRR}, abs/1708.02383, 2017.
\newblock URL \url{http://arxiv.org/abs/1708.02383}.

\bibitem[Franceschi et~al.(2018)Franceschi, Frasconi, Salzo, Grazzi, and
  Pontil]{franceschi2018bilevel}
Luca Franceschi, Paolo Frasconi, Saverio Salzo, Riccardo Grazzi, and
  Massimiliano Pontil.
\newblock Bilevel programming for hyperparameter optimization and
  meta-learning.
\newblock In \emph{International Conference on Machine Learning}, pages
  1568--1577. PMLR, 2018.

\bibitem[Gal et~al.(2017)Gal, Islam, and Ghahramani]{gal2017deep}
Yarin Gal, Riashat Islam, and Zoubin Ghahramani.
\newblock Deep bayesian active learning with image data.
\newblock In \emph{International Conference on Machine Learning}, pages
  1183--1192. PMLR, 2017.

\bibitem[Grazzi et~al.(2020)Grazzi, Franceschi, Pontil, and
  Salzo]{grazzi2020iteration}
Riccardo Grazzi, Luca Franceschi, Massimiliano Pontil, and Saverio Salzo.
\newblock On the iteration complexity of hypergradient computation.
\newblock 2020.

\bibitem[He et~al.(2016)He, Zhang, Ren, and Sun]{he2016deep}
Kaiming He, Xiangyu Zhang, Shaoqing Ren, and Jian Sun.
\newblock Deep residual learning for image recognition.
\newblock In \emph{Proceedings of the IEEE conference on computer vision and
  pattern recognition}, pages 770--778, 2016.

\bibitem[Houlsby et~al.(2011)Houlsby, Husz{\'a}r, Ghahramani, and
  Lengyel]{houlsby2011bayesian}
Neil Houlsby, Ferenc Husz{\'a}r, Zoubin Ghahramani, and M{\'a}t{\'e} Lengyel.
\newblock Bayesian active learning for classification and preference learning.
\newblock \emph{arXiv preprint arXiv:1112.5745}, 2011.

\bibitem[Ilyas et~al.(2022)Ilyas, Park, Engstrom, Leclerc, and
  Madry]{ilyas2022datamodels}
Andrew Ilyas, Sung~Min Park, Logan Engstrom, Guillaume Leclerc, and Aleksander
  Madry.
\newblock Datamodels: Understanding predictions with data and data with
  predictions.
\newblock In \emph{International Conference on Machine Learning}, pages
  9525--9587. PMLR, 2022.

\bibitem[Jackson et~al.(2019)Jackson, Presanis, Conti, and
  De~Angelis]{jackson2019value}
Christopher Jackson, Anne Presanis, Stefano Conti, and Daniela De~Angelis.
\newblock Value of information: Sensitivity analysis and research design in
  bayesian evidence synthesis.
\newblock \emph{Journal of the American Statistical Association}, 114\penalty0
  (528):\penalty0 1436--1449, 2019.

\bibitem[Jiang et~al.(2021)Jiang, Nagarajan, Baek, and
  Kolter]{jiang2021assessing}
Yiding Jiang, Vaishnavh Nagarajan, Christina Baek, and J~Zico Kolter.
\newblock Assessing generalization of sgd via disagreement.
\newblock In \emph{International Conference on Learning Representations}, 2021.

\bibitem[Just et~al.(2023)Just, Kang, Wang, Zeng, Ko, Jin, and
  Jia]{just2023lava}
Hoang~Anh Just, Feiyang Kang, Tianhao Wang, Yi~Zeng, Myeongseob Ko, Ming Jin,
  and Ruoxi Jia.
\newblock {LAVA}: Data valuation without pre-specified learning algorithms.
\newblock In \emph{The Eleventh International Conference on Learning
  Representations}, 2023.
\newblock URL \url{https://openreview.net/forum?id=JJuP86nBl4q}.

\bibitem[Karatzoglou et~al.(2013)Karatzoglou, Baltrunas, and
  Shi]{karatzoglou2013learning}
Alexandros Karatzoglou, Linas Baltrunas, and Yue Shi.
\newblock Learning to rank for recommender systems.
\newblock In \emph{Proceedings of the 7th ACM Conference on Recommender
  Systems}, pages 493--494, 2013.

\bibitem[Killamsetty et~al.(2021)Killamsetty, Sivasubramanian, Ramakrishnan,
  and Iyer]{killamsetty2021glister}
Krishnateja Killamsetty, Durga Sivasubramanian, Ganesh Ramakrishnan, and
  Rishabh Iyer.
\newblock Glister: Generalization based data subset selection for efficient and
  robust learning.
\newblock In \emph{Proceedings of the AAAI Conference on Artificial
  Intelligence}, volume~35, pages 8110--8118, 2021.

\bibitem[Ko et~al.(1995)Ko, Lee, and Queyranne]{ko1995exact}
Chun-Wa Ko, Jon Lee, and Maurice Queyranne.
\newblock An exact algorithm for maximum entropy sampling.
\newblock \emph{Operations Research}, 43\penalty0 (4):\penalty0 684--691, 1995.

\bibitem[Konyushkova et~al.(2017)Konyushkova, Sznitman, and
  Fua]{konyushkova2017learning}
Ksenia Konyushkova, Raphael Sznitman, and Pascal Fua.
\newblock Learning active learning from data.
\newblock \emph{Advances in neural information processing systems}, 30, 2017.

\bibitem[Kossen et~al.(2022)Kossen, Farquhar, Gal, and
  Rainforth]{kossen2022active}
Jannik Kossen, Sebastian Farquhar, Yarin Gal, and Thomas Rainforth.
\newblock Active surrogate estimators: An active learning approach to
  label-efficient model evaluation.
\newblock \emph{Advances in Neural Information Processing Systems},
  35:\penalty0 24557--24570, 2022.

\bibitem[Krizhevsky et~al.(2009)Krizhevsky, Hinton,
  et~al.]{krizhevsky2009learning}
Alex Krizhevsky, Geoffrey Hinton, et~al.
\newblock Learning multiple layers of features from tiny images.
\newblock 2009.

\bibitem[LeCun et~al.(1998)LeCun, Bottou, Bengio, and
  Haffner]{lecun1998gradient}
Yann LeCun, L{\'e}on Bottou, Yoshua Bengio, and Patrick Haffner.
\newblock Gradient-based learning applied to document recognition.
\newblock \emph{Proceedings of the IEEE}, 86\penalty0 (11):\penalty0
  2278--2324, 1998.

\bibitem[Lee et~al.(2019)Lee, Lee, Kim, Kosiorek, Choi, and Teh]{lee2019set}
Juho Lee, Yoonho Lee, Jungtaek Kim, Adam Kosiorek, Seungjin Choi, and Yee~Whye
  Teh.
\newblock Set transformer: A framework for attention-based
  permutation-invariant neural networks.
\newblock In \emph{International conference on machine learning}, pages
  3744--3753. PMLR, 2019.

\bibitem[Li(2022)]{li2022learning}
Hang Li.
\newblock \emph{Learning to rank for information retrieval and natural language
  processing}.
\newblock Springer Nature, 2022.

\bibitem[Li et~al.(2021)Li, Liu, van~de Weijer, and Raducanu]{li2021learning}
Minghan Li, Xialei Liu, Joost van~de Weijer, and Bogdan Raducanu.
\newblock Learning to rank for active learning: A listwise approach.
\newblock In \emph{2020 25th International Conference on Pattern Recognition
  (ICPR)}, pages 5587--5594. IEEE, 2021.

\bibitem[Li and Oliva(2021)]{li2021active}
Yang Li and Junier Oliva.
\newblock Active feature acquisition with generative surrogate models.
\newblock In \emph{International Conference on Machine Learning}, pages
  6450--6459. PMLR, 2021.

\bibitem[Lindley(1956)]{lindley1956measure}
Dennis~V Lindley.
\newblock On a measure of the information provided by an experiment.
\newblock \emph{The Annals of Mathematical Statistics}, 27\penalty0
  (4):\penalty0 986--1005, 1956.

\bibitem[Liu et~al.(2019)Liu, Davison, and Johns]{liu2019self}
Shikun Liu, Andrew Davison, and Edward Johns.
\newblock Self-supervised generalisation with meta auxiliary learning.
\newblock \emph{Advances in Neural Information Processing Systems}, 32, 2019.

\bibitem[Liu et~al.(2009)]{liu2009learning}
Tie-Yan Liu et~al.
\newblock Learning to rank for information retrieval.
\newblock \emph{Foundations and Trends{\textregistered} in Information
  Retrieval}, 3\penalty0 (3):\penalty0 225--331, 2009.

\bibitem[Maclaurin et~al.(2015)Maclaurin, Duvenaud, and
  Adams]{maclaurin2015gradient}
Dougal Maclaurin, David Duvenaud, and Ryan Adams.
\newblock Gradient-based hyperparameter optimization through reversible
  learning.
\newblock In \emph{International conference on machine learning}, pages
  2113--2122. PMLR, 2015.

\bibitem[Mussmann et~al.(2022)Mussmann, Reisler, Tsai, Mousavi, O'Brien, and
  Goldszmidt]{mussmann2022active}
Stephen Mussmann, Julia Reisler, Daniel Tsai, Ehsan Mousavi, Shayne O'Brien,
  and Moises Goldszmidt.
\newblock Active learning with expected error reduction.
\newblock \emph{arXiv preprint arXiv:2211.09283}, 2022.

\bibitem[Netzer et~al.(2011)Netzer, Wang, Coates, Bissacco, Wu, and
  Ng]{netzer2011reading}
Yuval Netzer, Tao Wang, Adam Coates, Alessandro Bissacco, Bo~Wu, and Andrew~Y
  Ng.
\newblock Reading digits in natural images with unsupervised feature learning.
\newblock 2011.

\bibitem[Parvaneh et~al.(2022)Parvaneh, Abbasnejad, Teney, Haffari, Van
  Den~Hengel, and Shi]{parvaneh2022active}
Amin Parvaneh, Ehsan Abbasnejad, Damien Teney, Gholamreza~Reza Haffari, Anton
  Van Den~Hengel, and Javen~Qinfeng Shi.
\newblock Active learning by feature mixing.
\newblock In \emph{Proceedings of the IEEE/CVF Conference on Computer Vision
  and Pattern Recognition}, pages 12237--12246, 2022.

\bibitem[Paszke et~al.(2017)Paszke, Gross, Chintala, Chanan, Yang, DeVito, Lin,
  Desmaison, Antiga, and Lerer]{paszke2017automatic}
Adam Paszke, Sam Gross, Soumith Chintala, Gregory Chanan, Edward Yang, Zachary
  DeVito, Zeming Lin, Alban Desmaison, Luca Antiga, and Adam Lerer.
\newblock Automatic differentiation in pytorch.
\newblock 2017.

\bibitem[Rajeswaran et~al.(2019)Rajeswaran, Finn, Kakade, and
  Levine]{rajeswaran2019meta}
Aravind Rajeswaran, Chelsea Finn, Sham~M Kakade, and Sergey Levine.
\newblock Meta-learning with implicit gradients.
\newblock \emph{Advances in neural information processing systems}, 32, 2019.

\bibitem[Roth and Small(2006)]{roth2006margin}
Dan Roth and Kevin Small.
\newblock Margin-based active learning for structured output spaces.
\newblock In \emph{Machine Learning: ECML 2006: 17th European Conference on
  Machine Learning Berlin, Germany, September 18-22, 2006 Proceedings 17},
  pages 413--424. Springer, 2006.

\bibitem[Roy and McCallum(2001)]{roy2001toward}
Nicholas Roy and Andrew McCallum.
\newblock Toward optimal active learning through monte carlo estimation of
  error reduction.
\newblock \emph{ICML, Williamstown}, 2:\penalty0 441--448, 2001.

\bibitem[Saran et~al.(2023)Saran, Yousefi, Krishnamurthy, Langford, and
  Ash]{saran2023streaming}
Akanksha Saran, Safoora Yousefi, Akshay Krishnamurthy, John Langford, and
  Jordan Ash.
\newblock Streaming active learning with deep neural networks.
\newblock In \emph{ICML 2023}, March 2023.

\bibitem[Saunshi et~al.(2022)Saunshi, Gupta, Braverman, and
  Arora]{saunshi2022understanding}
Nikunj Saunshi, Arushi Gupta, Mark Braverman, and Sanjeev Arora.
\newblock Understanding influence functions and datamodels via harmonic
  analysis.
\newblock \emph{arXiv preprint arXiv:2210.01072}, 2022.

\bibitem[Sener and Savarese(2017)]{sener2017active}
Ozan Sener and Silvio Savarese.
\newblock Active learning for convolutional neural networks: A core-set
  approach.
\newblock \emph{arXiv preprint arXiv:1708.00489}, 2017.

\bibitem[Settles(2012)]{settles2012active}
Burr Settles.
\newblock Active learning.
\newblock \emph{Synthesis lectures on artificial intelligence and machine
  learning}, 6\penalty0 (1):\penalty0 1--114, 2012.

\bibitem[Shen et~al.(2017)Shen, Yun, Lipton, Kronrod, and
  Anandkumar]{shen2017deep}
Yanyao Shen, Hyokun Yun, Zachary~C Lipton, Yakov Kronrod, and Animashree
  Anandkumar.
\newblock Deep active learning for named entity recognition.
\newblock \emph{arXiv preprint arXiv:1707.05928}, 2017.

\bibitem[Sinha et~al.(2019)Sinha, Ebrahimi, and Darrell]{sinha2019variational}
Samarth Sinha, Sayna Ebrahimi, and Trevor Darrell.
\newblock Variational adversarial active learning.
\newblock In \emph{Proceedings of the IEEE/CVF International Conference on
  Computer Vision}, pages 5972--5981, 2019.

\bibitem[Sourati et~al.(2016)Sourati, Akcakaya, Dy, Leen, and
  Erdogmus]{sourati2016classification}
Jamshid Sourati, Murat Akcakaya, Jennifer~G Dy, Todd~K Leen, and Deniz
  Erdogmus.
\newblock Classification active learning based on mutual information.
\newblock \emph{Entropy}, 18\penalty0 (2):\penalty0 51, 2016.

\bibitem[Team(2023)]{decileteam_2023_distil}
Decile Team.
\newblock distil.
\newblock \url{https://github.com/decile-team/distil}, 2023.

\bibitem[Verma et~al.(2022)Verma, Kawaguchi, Lamb, Kannala, Solin, Bengio, and
  Lopez-Paz]{verma2022interpolation}
Vikas Verma, Kenji Kawaguchi, Alex Lamb, Juho Kannala, Arno Solin, Yoshua
  Bengio, and David Lopez-Paz.
\newblock Interpolation consistency training for semi-supervised learning.
\newblock \emph{Neural Networks}, 145:\penalty0 90--106, 2022.

\bibitem[Wang et~al.(2021)Wang, Chen, and Jia]{oneroundactivelearn}
Tianhao Wang, Si~Chen, and Ruoxi Jia.
\newblock One-round active learning.
\newblock \emph{CoRR}, abs/2104.11843, 2021.
\newblock URL \url{https://arxiv.org/abs/2104.11843}.

\bibitem[Xiao et~al.(2017)Xiao, Rasul, and Vollgraf]{xiao2017fashion}
Han Xiao, Kashif Rasul, and Roland Vollgraf.
\newblock Fashion-mnist: a novel image dataset for benchmarking machine
  learning algorithms.
\newblock \emph{arXiv preprint arXiv:1708.07747}, 2017.

\bibitem[Xie et~al.(2022)Xie, Yuan, Li, Liu, and Cheng]{xie2022towards}
Binhui Xie, Longhui Yuan, Shuang Li, Chi~Harold Liu, and Xinjing Cheng.
\newblock Towards fewer annotations: Active learning via region impurity and
  prediction uncertainty for domain adaptive semantic segmentation.
\newblock In \emph{Proceedings of the IEEE/CVF Conference on Computer Vision
  and Pattern Recognition}, pages 8068--8078, 2022.

\bibitem[Yan et~al.(2022)Yan, Zhang, and He]{yan2022budget}
Shipeng Yan, Songyang Zhang, and Xuming He.
\newblock Budget-aware few-shot learning via graph convolutional network.
\newblock \emph{arXiv preprint arXiv:2201.02304}, 2022.

\bibitem[Yang et~al.(2019)Yang, Wu, and Arnold]{yang2019machine}
Kevin~K Yang, Zachary Wu, and Frances~H Arnold.
\newblock Machine-learning-guided directed evolution for protein engineering.
\newblock \emph{Nature methods}, 16\penalty0 (8):\penalty0 687--694, 2019.

\bibitem[Yehuda et~al.(2022)Yehuda, Dekel, Hacohen, and
  Weinshall]{yehuda2022active}
Ofer Yehuda, Avihu Dekel, Guy Hacohen, and Daphna Weinshall.
\newblock Active learning through a covering lens.
\newblock \emph{arXiv preprint arXiv:2205.11320}, 2022.

\bibitem[Yoo and Kweon(2019)]{yoo2019learning}
Donggeun Yoo and In~So Kweon.
\newblock Learning loss for active learning.
\newblock In \emph{Proceedings of the IEEE/CVF conference on computer vision
  and pattern recognition}, pages 93--102, 2019.

\bibitem[Zaheer et~al.(2017)Zaheer, Kottur, Ravanbakhsh, Poczos, Salakhutdinov,
  and Smola]{zaheer2017deep}
Manzil Zaheer, Satwik Kottur, Siamak Ravanbakhsh, Barnabas Poczos, Russ~R
  Salakhutdinov, and Alexander~J Smola.
\newblock Deep sets.
\newblock \emph{Advances in neural information processing systems}, 30, 2017.

\bibitem[Zhong et~al.(2021)Zhong, Ghosh, Klein, and
  Steinhardt]{zhong2021larger}
Ruiqi Zhong, Dhruba Ghosh, Dan Klein, and Jacob Steinhardt.
\newblock Are larger pretrained language models uniformly better? comparing
  performance at the instance level.
\newblock In \emph{Findings of the Association for Computational Linguistics:
  ACL-IJCNLP 2021}, pages 3813--3827, 2021.

\end{thebibliography}

\appendix
\section{Appendix}

\begin{figure*}[!t]
\label{SizeofSeedSet}
\centering
\begin{subfigure}{.33\textwidth}
        \centering
\includegraphics[clip,trim=0cm 0cm 0cm 0cm,width=\textwidth]{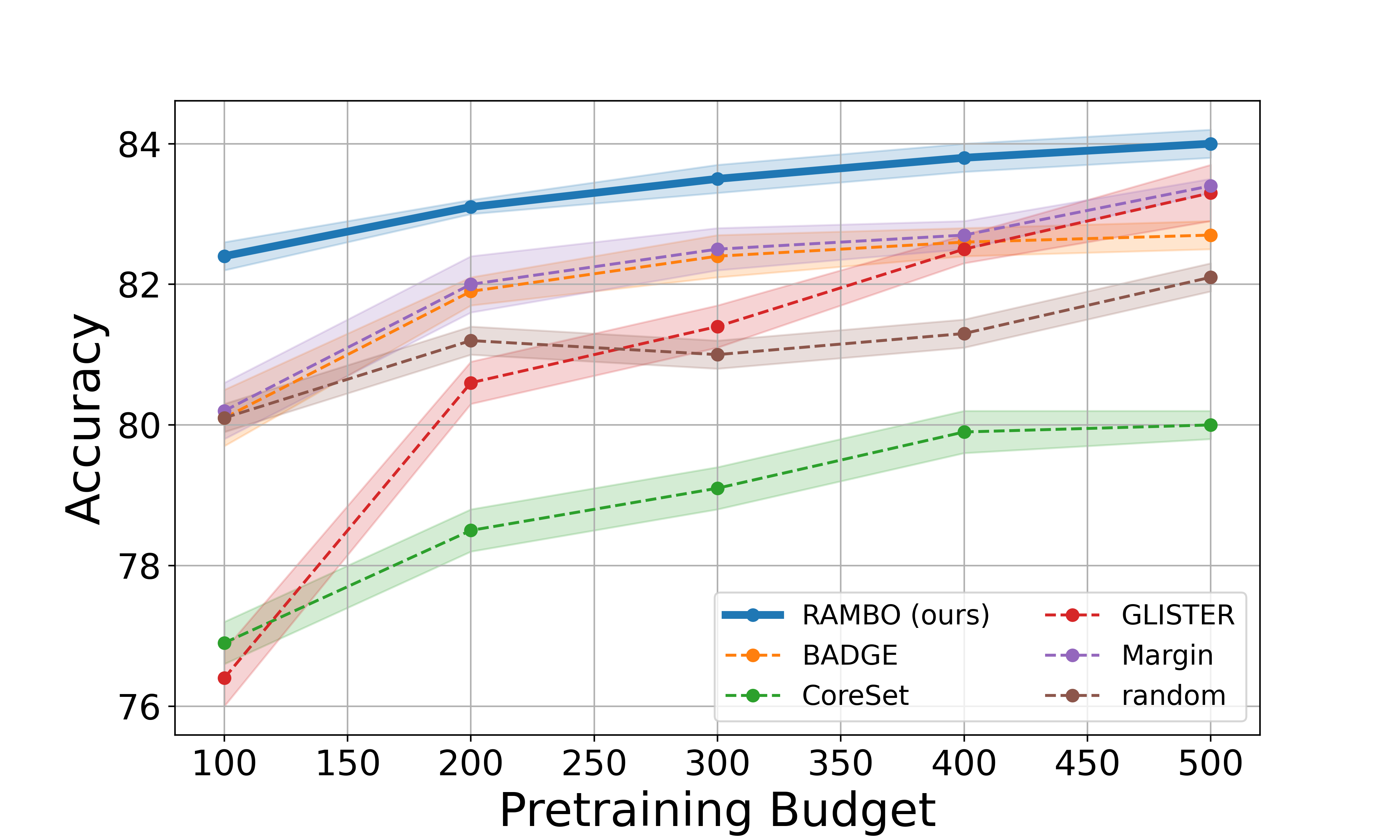}
    \caption{\footnotesize FashionMNIST}
    \label{Pretraining Budget Variation FashionMNIST}
     \end{subfigure}\hfil
    \begin{subfigure}{.33\textwidth}
        \centering
        \includegraphics[clip,trim=0cm 0cm 0cm 0cm,width=\textwidth]{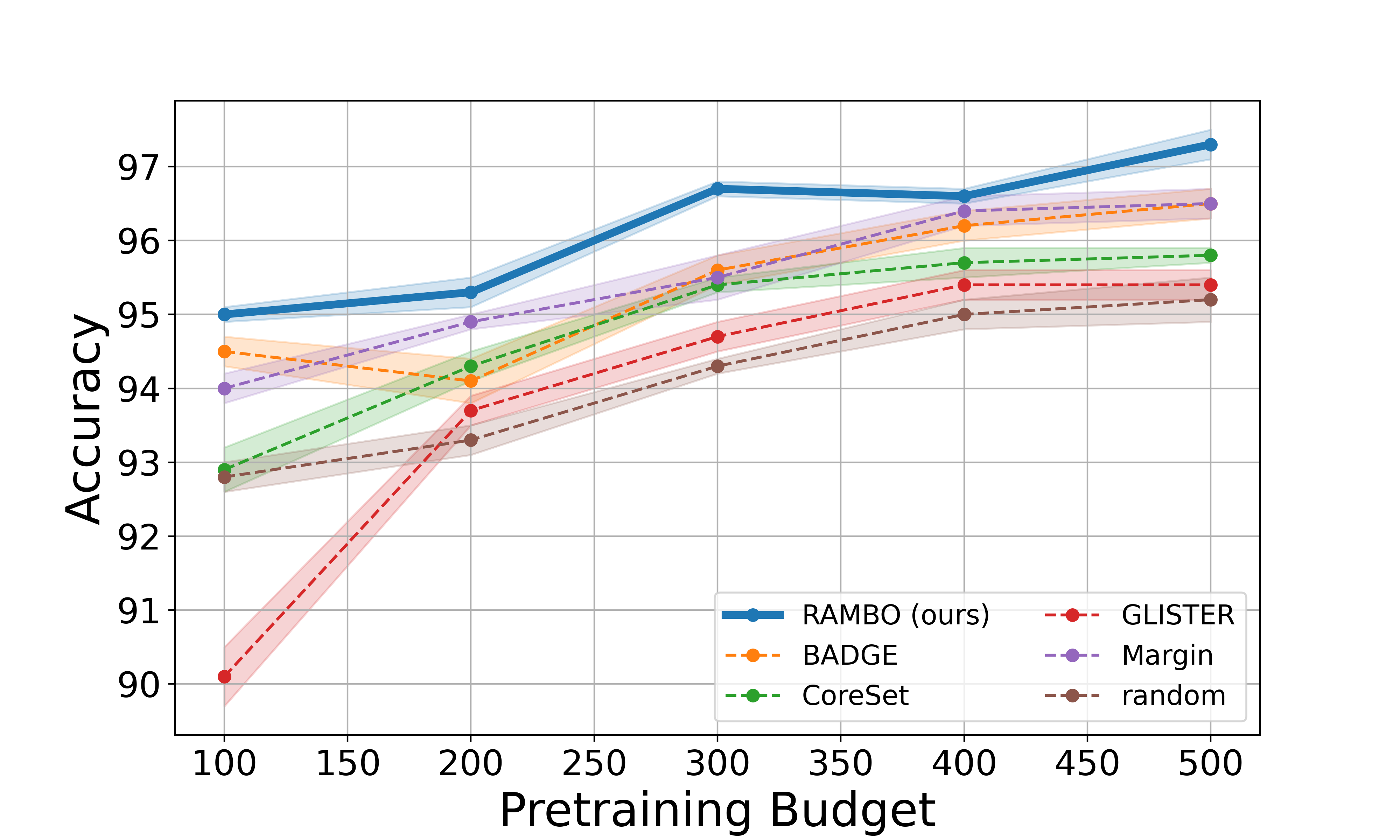}
        \caption{MNIST}\label{}
    \end{subfigure}\hfil
    \begin{subfigure}{.33\textwidth}
        \centering
        \includegraphics[clip,trim=0cm 0cm 0cm 0cm,width=\textwidth]{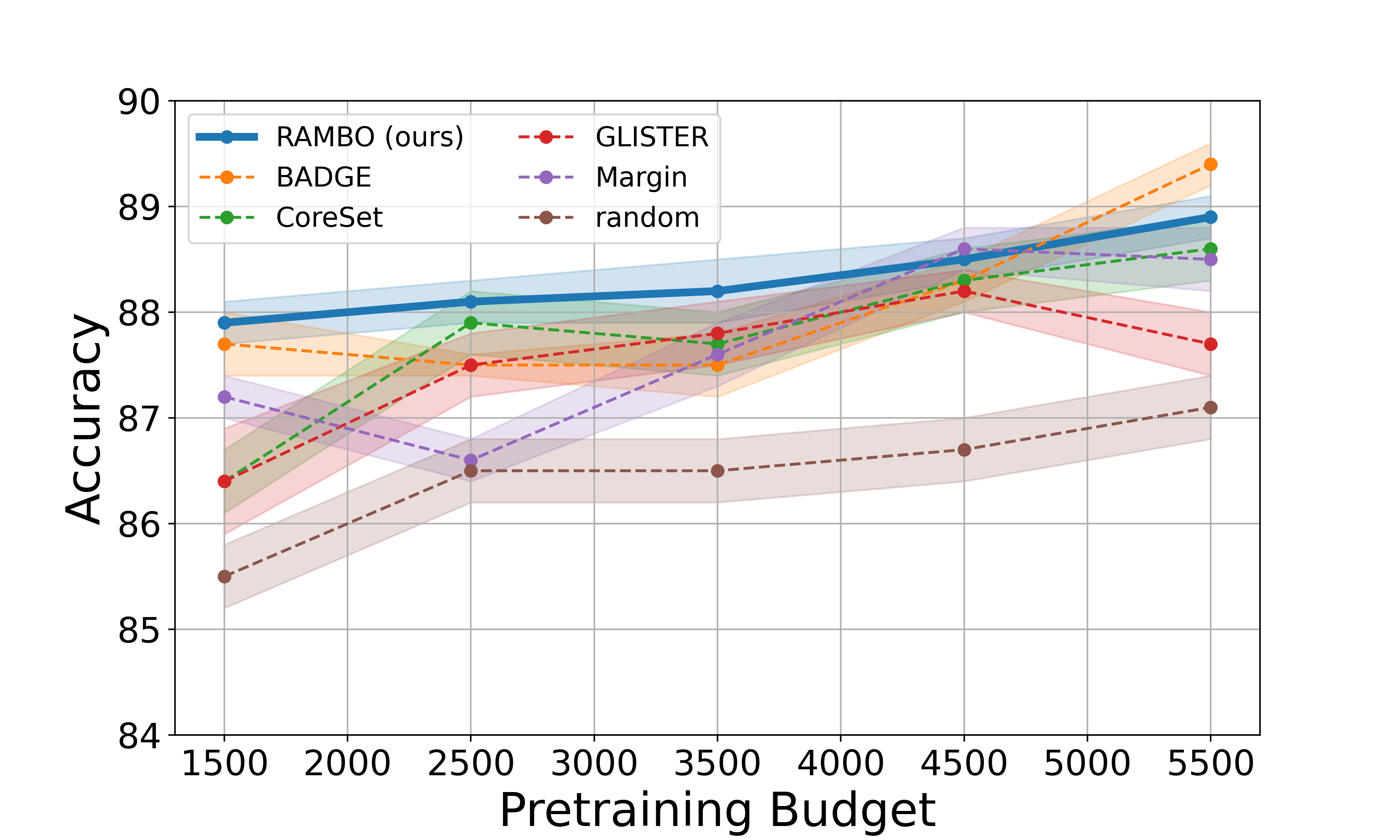}
        \caption{SVHN}\label{}
    \end{subfigure}
    \caption{Experimental results. \textbf{(a-c)} Active learning validation performance with $B = 500$ for FashionMNIST and MNIST and $B = 5000$ for SVHN. Results are given in \%.} 
    \label{SizeofSeedSet}
\end{figure*}

\subsection{Utility Model Architecture}
\label{UtilityModelArchitecture}
Here, we describe the acquisition network (Utility Model) discussed in Section \ref{experimental_setup}. 
\subsubsection{MNIST and FashionMNIST}
\label{FeatureExtractorMNIST&FashionMNIST}
Our architecture performs the following operations on pairs of subsets of images (Utility Samples) with equal size. We use below networks as feature extractor for pairs of raw embeddings of images. For each one within the pair:

1. 2-D convolution on set of images.

2. 2-D Max Pool on output of (1).

3. ReLU on output of (2).

4. 2-D DropOut on output of (3).

5. 2-D Max Pool on output of (4).

6. ReLU on output of (5).

7. Fully-Connected Layer on Output of (6).

8. ReLU on output of (7).

9. 2-D DropOut on output of (8).

10. Fully-Connected Layer on output of (9).

11. ReLU on output of (10).

\subsubsection{CIFAR10 and SVHN}
\label{FeatureExtractorforCIFAR10&SVHN}
We use pretrained ResNet-18 on ImageNet as feature extractor and perform the following operations on pairs of subsets of extracted features for each image. For each one within the pair: 

1. Fully Connected Layer on set of feature embeddings.

2. ReLU on output of (1).

3. Fully Connected Layer on output of (2).

\subsubsection{Mutitask Set-based Neural Networks with RankNet}
\label{multitask_description}

After average pooling of output of (11) for MNIST and FashionMNIST and output of (3) for CIFAR10 and SVHN, for each one within the pair, we perform the following operations:

1. Fully Connected Layer on extracted features

2. ReLU on output of (1).

3. Fully Connected Layer on output of (2).

4. Sigmoid function on output of (3). 

Denote the output of (4) as $\phi_{1}$ and $\phi_{2}$. 

For the prediction of probability score that which subset has larger utility value in the pair, we apply RankNet on $\phi_{1}$ and $\phi_{2}$ for pair comparison. The output score predicted by RankNet is the final probability score that we shall use to determine whether the first set has larger utility value than the second.

For the interpolation of utility value, we use $\phi_{1}$ and $\phi_{2}$ as embedding.

For the prediction of optimal transport distance, we use MLP projection head for $\phi_{1}$ and $\phi_{2}$:

1. Fully Connected Layer on $\phi_{1}$ and $\phi_{2}$

2. ReLU on outputs of (1)

3. Fully-Connected Layer on outputs of (2).

We use the outputs of (3) as a supervision signal in designing the loss function for the neural acquisition function (see Definition \ref{otloss} in Section \ref{DUAL_MAX}).

We chose $\lambda_{1}, \lambda_{2}$ to be $0.5$ and $\lambda_{3}$ to be 1.


\subsection{Supplemental Experimental Results}
In the main text, we have focused our evaluation on CIFAR-10. Here, we provide experiments to show effectiveness of \algname on diverse datasets such as MNIST, FashionMNIST and SVHN for single round unlabeled data selection. In the main paper, we fix $k = 200$ for FashionMNIST and MNIST and $k = 2500$ for CIFAR10 and SVHN in  Figure~\ref{Accuracy Validation Performance}(a-d). We construct all the pretraining set by random sampling from the full training set of each dataset.

\subsubsection{Size of Pretraining Set}
Figure~\ref{SizeofSeedSet} illustrates the impact of the size of pretraining set on final validation set accuracy. One shall see \algname outperforms the rest of baselines with most of pretraining splits. The only outlier case could be SVHN, similar to CIFAR10 setting in the main paper where $k = 5500$ and $B=5000$. One possibility could be $k = 5500$ is suffice for BADGE to learn an accurate-enough gradient embedding space for single round selection. Another interesting observation is GLISTER often performs worse than most of baselines for FashionMNIST and MNIST when pretraining budget is extremely low as $k = 100$. One possible explanation could be extremely small pretraining budget can not guarantee good inner-level optimization for maximizing training set log-likelihood for extremely small labeled data.

\subsubsection{Bi-level training, OT Distance and RankNet}
For simplicity, the checkmarks for optimal transport means $\lambda_{OT} = 1$ and the crossmarks for RankNet denotes regression-based utility model as stated in the main paper. In particular, we only collect single utility sample and develop multitask learning framework on the single utility sample. We still use the feature extractor explained in Section~\ref{FeatureExtractorMNIST&FashionMNIST} for MNIST and FashionMNIST and Section~\ref{FeatureExtractorforCIFAR10&SVHN} for CIFAR10 and SVHN. For the regression style acquisition function, we impose MLP head on the shared representation space $\phi$ for predicting validation accuracy with $\hat{u} = g(\phi) = W^{(2)}(\sigma(W^{(1)}))$ where $\sigma$ is a RELU non-linearity, very much similar to the description of predicting OT distance in Section~\ref{multitask_description}. For OT distance regularization, we adopt the same MLP projection head architecture described in Section~\ref{multitask_description}.

To prove the efficacy of synergizing three seemingly irrelevant submodules together, we provide ablation study of three submodules for the rest of three datasets. Table~\ref{BilevelTrainingFashionMNIST}, ~\ref{BilevelTrainingMNIST} and \ref{BilevelTrainingSVHN} show the impact of turning off each submodule on the final validation set accuracy for FashionMNIST, MNIST and SVHN respectively.


\begin{table}[!t]
\centering
\small
\caption{Ablation study on three submodules with pretraining set $k=200$ and acquisition budget $B=500$ for FashionMNIST. The last row 
corresponds to the random baseline.}
\label{BilevelTrainingFashionMNIST}
\scalebox{1.00}{
\begin{tabular}{lccc}
    \toprule
    Bilevel & Optimal Transport & RankNet & Accuracy\\
    \midrule
     $\checkmark$ & $\checkmark$ & $\checkmark$ & $\mathbf{83.1 \pm 0.1}$ \\
    $\checkmark$ & $\checkmark$ & $\times$ & $81.9 \pm 0.2$\\
    $\checkmark$ & $\times$ & $\checkmark$ & $81.2 \pm 0.4$\\
    $\checkmark$ & $\times$ & $\times$ & $81.8 \pm 0.2$ \\
    $\times$ & $\checkmark$ & $\checkmark$ & $81.0 \pm 0.3$\\
     $\times$ & $\checkmark$ & $\times$ & $81.7 \pm 0.2$ \\
    $\times$ & $\times$ & $\checkmark$ & $80.9 \pm 0.3$ \\
    $\times$ & $\times $ & $\times$ & $81.6 \pm 0.1$ \\
    - & - & - & $81.2 \pm 0.2$ \\
    \bottomrule
\end{tabular}}
\end{table}

\begin{table}[!t]
\centering
\small
\caption{Ablation study on three submodules with pretraining set $k=200$ and acquisition budget $B=500$ for MNIST. The last row 
corresponds to the random baseline.}
\label{BilevelTrainingMNIST}
\scalebox{1.00}{
\begin{tabular}{lccc}
    \toprule
    Bilevel & Optimal Transport & RankNet & Accuracy\\
    \midrule
     $\checkmark$ & $\checkmark$ & $\checkmark$ & $\mathbf{95.3 \pm 0.2}$ \\
    $\checkmark$ & $\checkmark$ & $\times$ & $94.9 \pm 0.2$\\
    $\checkmark$ & $\times$ & $\checkmark$ & $95.0 \pm 0.1$\\
    $\checkmark$ & $\times$ & $\times$ & $94.8 \pm 0.2$ \\
    $\times$ & $\checkmark$ & $\checkmark$ & $94.6 \pm 0.1$\\
     $\times$ & $\checkmark$ & $\times$ & $94.9 \pm 0.1$ \\
    $\times$ & $\times$ & $\checkmark$ & $95.0 \pm 0.2$ \\
    $\times$ & $\times $ & $\times$ & $94.8 \pm 0.2$ \\
    - & - & - & $93.4 \pm 0.1$ \\
    \bottomrule
\end{tabular}}
\end{table}

\begin{table}[!t]
\centering
\small
\caption{Ablation study on three submodules with pretraining set $k=3500$ and acquisition budget $B=5000$ for SVHN. The last row 
corresponds to the random baseline.}
\label{BilevelTrainingSVHN}
\scalebox{1.00}{
\begin{tabular}{lccc}
    \toprule
    Bilevel & Optimal Transport & RankNet & Accuracy\\
    \midrule
     $\checkmark$ & $\checkmark$ & $\checkmark$ & $\mathbf{88.1 \pm 0.3}$ \\
    $\checkmark$ & $\checkmark$ & $\times$ & $86.7 \pm 0.2$\\
    $\checkmark$ & $\times$ & $\checkmark$ & $87.8 \pm 0.3$ \\
    $\checkmark$ & $\times$ & $\times$ & $86.5 \pm 0.3$ \\
    $\times$ & $\checkmark$ & $\checkmark$ & $86.1 \pm 0.2$\\
     $\times$ & $\checkmark$ & $\times$ & $87.8 \pm 0.2$ \\
    $\times$ & $\times$ & $\checkmark$ & $87.5 \pm 0.1$ \\
    $\times$ & $\times $ & $\times$ & $86.1 \pm 0.2$ \\
    - & - & - & $86.5 \pm 0.3$ \\
    \bottomrule
\end{tabular}}
\end{table}

\begin{figure*}[!t]
    \begin{subfigure}{.33\textwidth}
        \centering
\includegraphics[clip,trim=0cm 0cm 0cm 0cm,width=\textwidth]{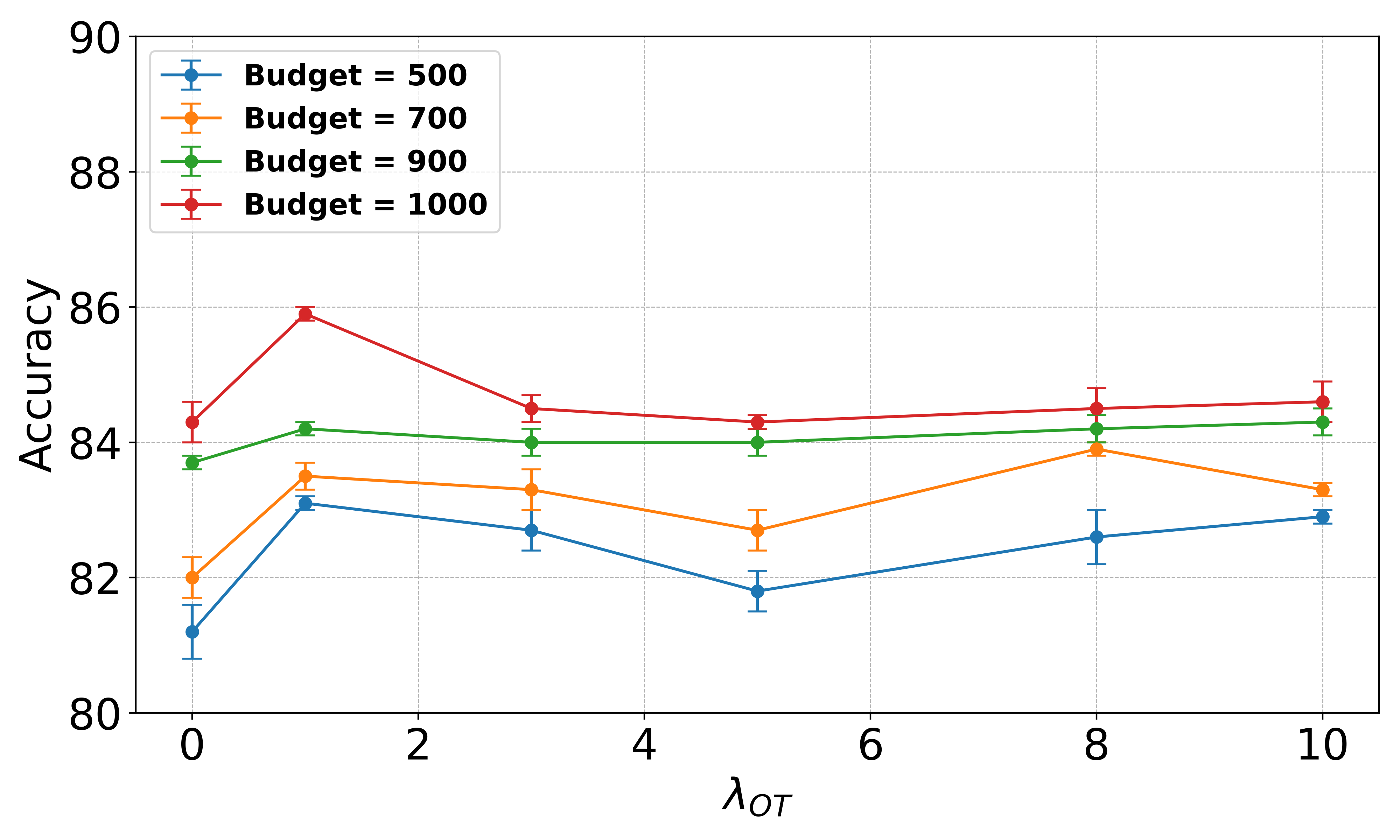}
    \caption{\footnotesize FashionMNIST}
    \label{Pretraining Budget Variation FashionMNIST}
     \end{subfigure}
    \begin{subfigure}{.33\textwidth}
        \centering
        \includegraphics[clip,trim=0cm 0cm 0cm 0cm,width=\textwidth]{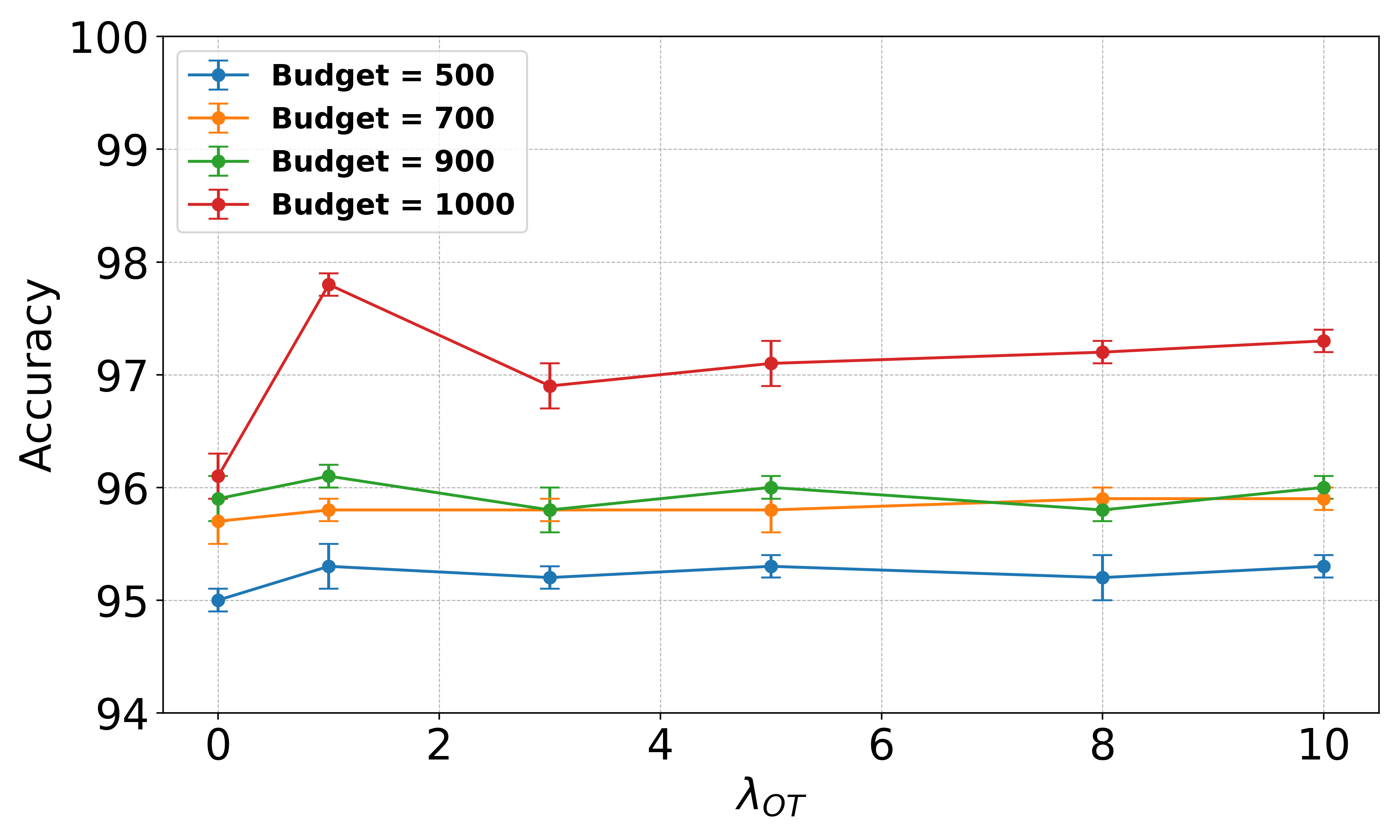}
        \caption{MNIST}\label{}
    \end{subfigure}\hfil
    \begin{subfigure}{.33\textwidth}
        \centering
        \includegraphics[clip,trim=0cm 0cm 0cm 0cm,width=\textwidth]{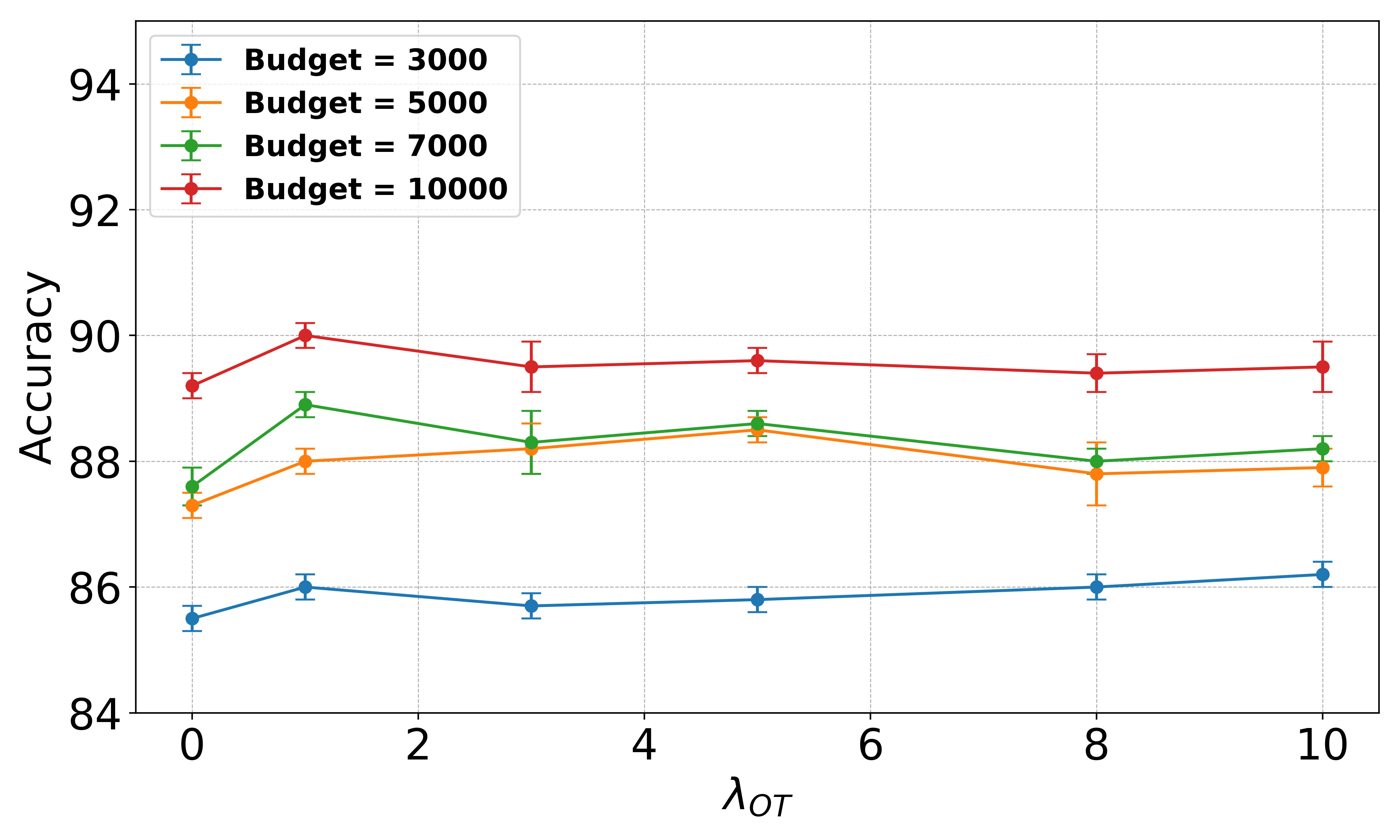}
        \caption{SVHN}\label{}
    \end{subfigure}\hfil
    \caption{Different choices of $\lambda_{OT}$ for pretraining set size $k = 200$ for (a) and (b) and $k = 2500$ for (c) by different acquisition budget} 
    \label{hyperparam_ot}
\end{figure*}

\subsubsection{Hyperparameter Tuning for OT distance}
Figure~\ref{Accuracy Validation Performance}(f) illustrates the benefits of incorporating optimal transport distance into the loss structure of our utility model. Figure~\ref{hyperparam_ot} shall serve as a complement to uncover the usefulness of optimal transport distance, regardless of the scale of $\lambda_{OT}$, for various datasets of interest. Regardless of datasets and classification networks architecture, the incorporation of optimal transport distance finds utility in reducing generalization error, measured by the increase of validation set accuracy. Even though $\lambda_{OT}$ can be a hard hyperparameter for fine-tuning, either Figure~\ref{Accuracy Validation Performance}(f) and Figure~\ref{hyperparam_ot} suggest final validation set accuracy for $\lambda_{OT} \neq 0$ is higher than its counterpart for $\lambda_{OT} = 0$.

\subsubsection{Runtime 
Analysis}
All models are trained using NVIDIA A40 GPU with 48GB. No Parallelism and As stated in main text, all the experiments are repeated for 10 trials. We fix $k = 2500$ and $B = 5000$ for CIFAR10 and SVHN with $n = 30$ utility samples collected per batch with $\tau_{1} = 2$, $b = 1000$ and $k_{1} = 500$ for pretraining stage with each batch trained for 20 epochs. For CIFAR10, the total training time for both pretraining and acquisition stage is 1 hour and 20 minutes with pretraining stage 40 minutes and acquisition stage 20 minutes. For SVHN, the total training time for both pretraining and acquisition stage is roughly 1 hour with pretraining stage 29 minutes and acquisition stage 34 minutes.

We fix $k = 200$ and $B = 500$ for MNIST and FashionMNIST with $n = 50$ utility samples collected per batch with $\tau_{1} = 3$, $b = 50$ and $k_{1} = 50$ for pretraining stage with each batch trained for 20 epochs. For MNIST, the total training time for both pretraining and acquisition stage is 59 minutes with pretraining stage 39 minutes and acquisition stage 20 minutes. For FashionMNIST, the total training time for both pretraining and acquisition stage is 50 minutes with pretraining stage 32 minutes and acquisition stage 18 minutes.

\end{document}